\documentclass{article}

\PassOptionsToPackage{numbers, compress}{natbib}
\usepackage[preprint]{neurips_2019}

\usepackage[utf8]{inputenc} 
\usepackage[T1]{fontenc}    
\usepackage{hyperref}       
\usepackage{url}            
\usepackage{booktabs}       
\usepackage{amsfonts}       
\usepackage{nicefrac}       
\usepackage{microtype}      
\usepackage{amsmath}        
\usepackage{amssymb}        
\usepackage{acronym}        
\usepackage{todonotes}      
\usepackage{graphicx,subfigure} 
\usepackage{wrapfig}        
\usepackage{cleveref}       
\usepackage{algorithm}
\usepackage[noend]{algpseudocode}
\usepackage[toc,page]{appendix}

\acrodef{ODE}{Ordinary Differential Equation}
\acrodef{IVP}{Initial Value Problem}
\acrodef{VF}{Vector Field}

\acrodef{RK4}{Runge-Kutta 4}
\acrodef{DP}{Dormand-Prince}

\acrodef{NODE}[NODE]{Neural ODE}
\acrodef{ANODE}[ANODE]{Augmented NODE}
\acrodef{NFE}{Number of Function Evaluations}
\acrodef{MFE}[Max-NFE]{Maximal NFE}

\acrodef{SVF}{Stochastic VF}
\acrodef{VFM}{VF Mixture}
\acrodef{SVFM}{Stochastic VF Mixture}
\acrodef{TL}[TLoss]{Transportation Loss}
\acrodef{DV}[VLoss]{Variance Loss}
\acrodef{APL}[FLoss]{Forecasting Loss}
\acrodef{MDL}[MDLoss]{Mixture Density Loss}
\acrodef{PLL}[TVLoss]{Transport \& Variance Loss}

\acrodef{NN}{Neural Network}
\acrodef{RNN}{Recurrent NN}
\acrodef{CNN}{Convolutional NN}
\acrodef{RESNET}[ResNet]{Residual Network}

\acrodef{CRF}{Conditional Random Field}
\acrodef{HMM}{Hidden Markov Model}
\acrodef{EWMA}{Exponentially Weighted Moving Averaging}
\acrodef{TOD}[ToD]{Time of Day}
\acrodef{PDF}{Probability Density Function}
\acrodef{LIDAR}[LiDAR]{Light Detection And Ranging}
\acrodef{SLAM}{Simultaneous Localisation And Mapping}
\acrodef{XOR}{Exclusive-Or}        

\newcommand{\X}{\mathbf{X}}

\newcommand{\bfk}{\mathbf{k}}

\newcommand{\z}{\mathbf{z}}
\newcommand{\x}{\mathbf{x}}
\newcommand{\h}{\mathbf{h}}

\renewcommand{\u}{\mathbf{u}}

\newcommand{\thetab}{\boldsymbol{\theta}}

\newcommand{\pib}{\boldsymbol{\pi}}
\newcommand{\psib}{\boldsymbol{\psi}}
\newcommand{\Psib}{\boldsymbol{\Psi}}

\newcommand{\taub}{\boldsymbol{\tau}}

\newcommand{\mub}{\boldsymbol{\mu}}

\newcommand{\ie}{{\em i.e.~\/}}
\newcommand{\eg}{{\em e.g.~\/}}

\newcommand{\vs}{{\em vs.~\/}}
\newcommand{\cf}{{\em c.f.~\/}}

\renewcommand{\lim}{\operatornamewithlimits{lim}}

\DeclareMathOperator{\interp}{interp}

\makeatletter

\title{Neural ODEs with stochastic vector field mixtures}

\author{%
  Niall Twomey, Micha\l{} Koz\l{}owski, Ra\'ul Santos-Rodr\'iguez\\
  University of Bristol\\
  Bristol, United Kingdom \\
  \texttt{\{niall.twomey, m.kozlowski, enrsr\}@bristol.ac.uk} 
}

\begin{document}

\maketitle


\begin{abstract}



It was recently shown that neural ordinary differential equation models cannot solve fundamental and seemingly straightforward tasks even with high-capacity vector field representations. This paper introduces two other fundamental tasks to the set that baseline methods cannot solve, and proposes mixtures of stochastic vector fields as a model class that is capable of solving these essential problems. Dynamic vector field selection is of critical importance for our model, and our approach is to propagate component uncertainty over the integration interval with a technique based on forward filtering. We also formalise several loss functions that encourage desirable properties on the trajectory paths, and of particular interest are those that directly encourage fewer expected function evaluations. Experimentally, we demonstrate that our model class is capable of capturing the natural dynamics of human behaviour; a notoriously volatile application area. Baseline approaches cannot adequately model this problem. 


\end{abstract}

\section{Introduction}
\label{section:introduction}

Recent work has linked \acp{RESNET} \cite{he2016deep} to \acp{ODE} and demonstrated that \acp{RESNET} may be interpreted as Euler solutions to \acp{IVP} \cite{lu2017beyond,weinan2017proposal,ruthotto2018deep}. This idea has been further developed with \acp{RESNET} taken to their continuous limit by \citet{chen2018neural} with the introduction of \acp{NODE}. In their setting, \acp{NN} produce \ac{VF} representations that are utilised by \ac{ODE} solvers to learn and calculate explicit flow trajectories through the feature space. The resulting model can be seen to be of continuous depth since the \ac{ODE} solver will query the \acp{VF} at arbitrary levels that are unspecified in advance. Their work also demonstrates efficient and scalable inference with the adjoint trick.

Vector fields are defined as follows in the \ac{NODE} paradigm

\begin{align}
    \label{eq:node}
    \nabla \h(t_i) = f(\h(t_i), t_i; \thetab)
\end{align}

\noindent where $\h(t_i)\in\mathbb{R}^D$ is a hidden state at depth $t_i$ (base state $\h(t_0) \triangleq \x$, maximal depth $t_T$), $f$ is a neural network of arbitrary architecture and $\thetab$ are its parameters. Subsequent hidden states are derived by taking small steps from $\h(t_i)$ in the direction of the vector field, \ie $\h(t_{i+1}) = \h(t_i) + \delta_{t_i} \nabla \h(t_i)$, and the step size $\delta_{t_i}$ is often selected automatically by an \ac{ODE} solver. The solution of this \ac{IVP} is obtained by repeatedly taking these steps until the output state $\h(t_T)$ is reached, and these outputs may undergo a final transformation, \eg through a softmax layer in classification tasks. The optimisation objective is to adjust the dynamics of the \ac{ODE} through $\thetab$ to maximise the data likelihood. 

A critical component of \acp{NODE} is the parametrisation of the \acp{VF} and, since $f$ is a neural network, the practitioner has the important responsibility of appropriately specifying its architecture. Even though the \ac{VF} representations will be non-linear (and arbitrarily `flexible' depending on architecture choice), \ac{NODE} models cannot model arbitrary problems. This was clearly demonstrated by \citet{dupont2019augmented} with a seemingly straightforward task in one dimension. 
\begin{wrapfigure}{r}{0.42\textwidth}
    \centering
    \vspace{-1em}
    \subfigure[Crossing]{\label{fig:fail:cross:o}\includegraphics[height=22.5mm]{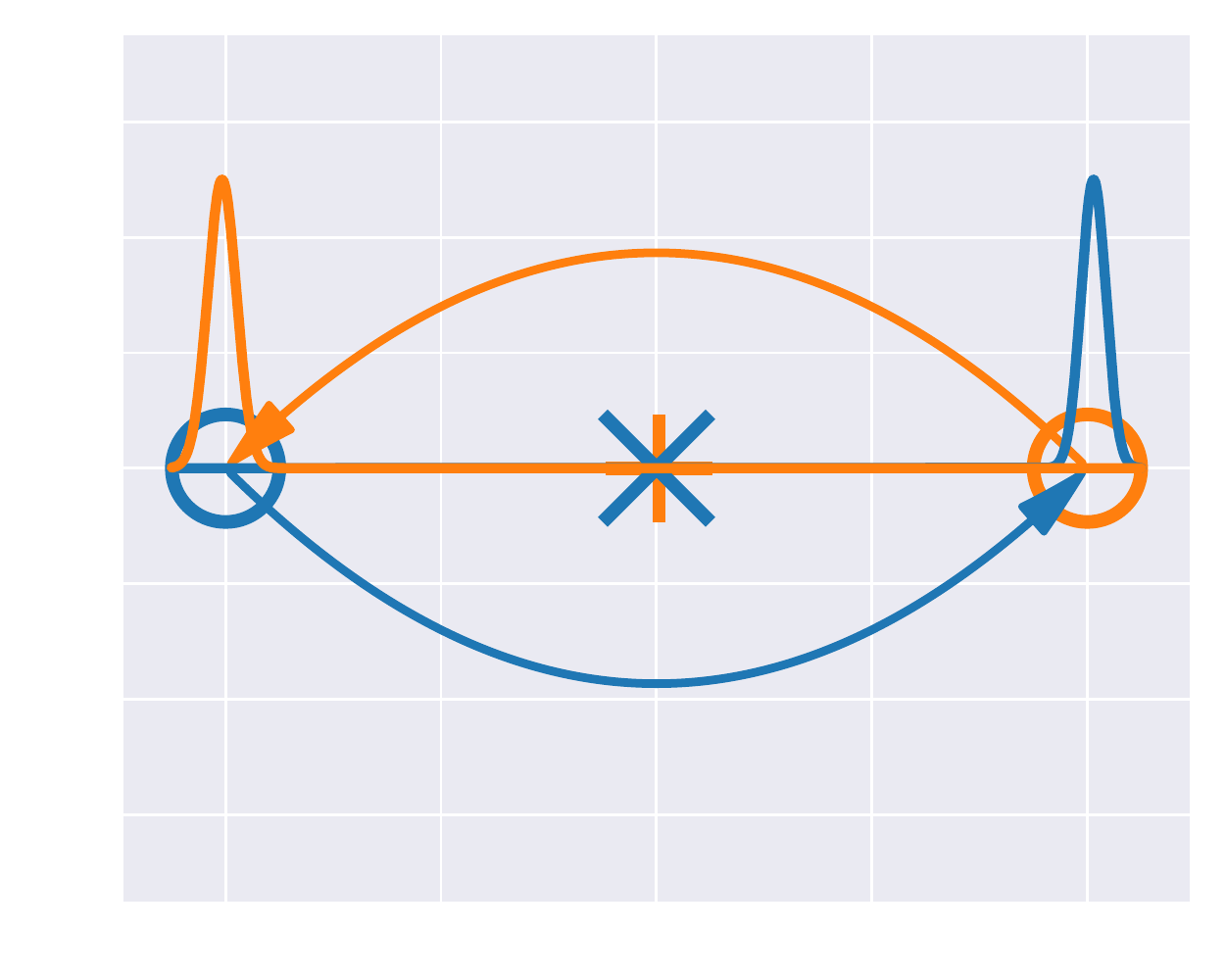}}
    \subfigure[Crossing (t)]{\label{fig:fail:cross:t}\includegraphics[height=22.5mm]{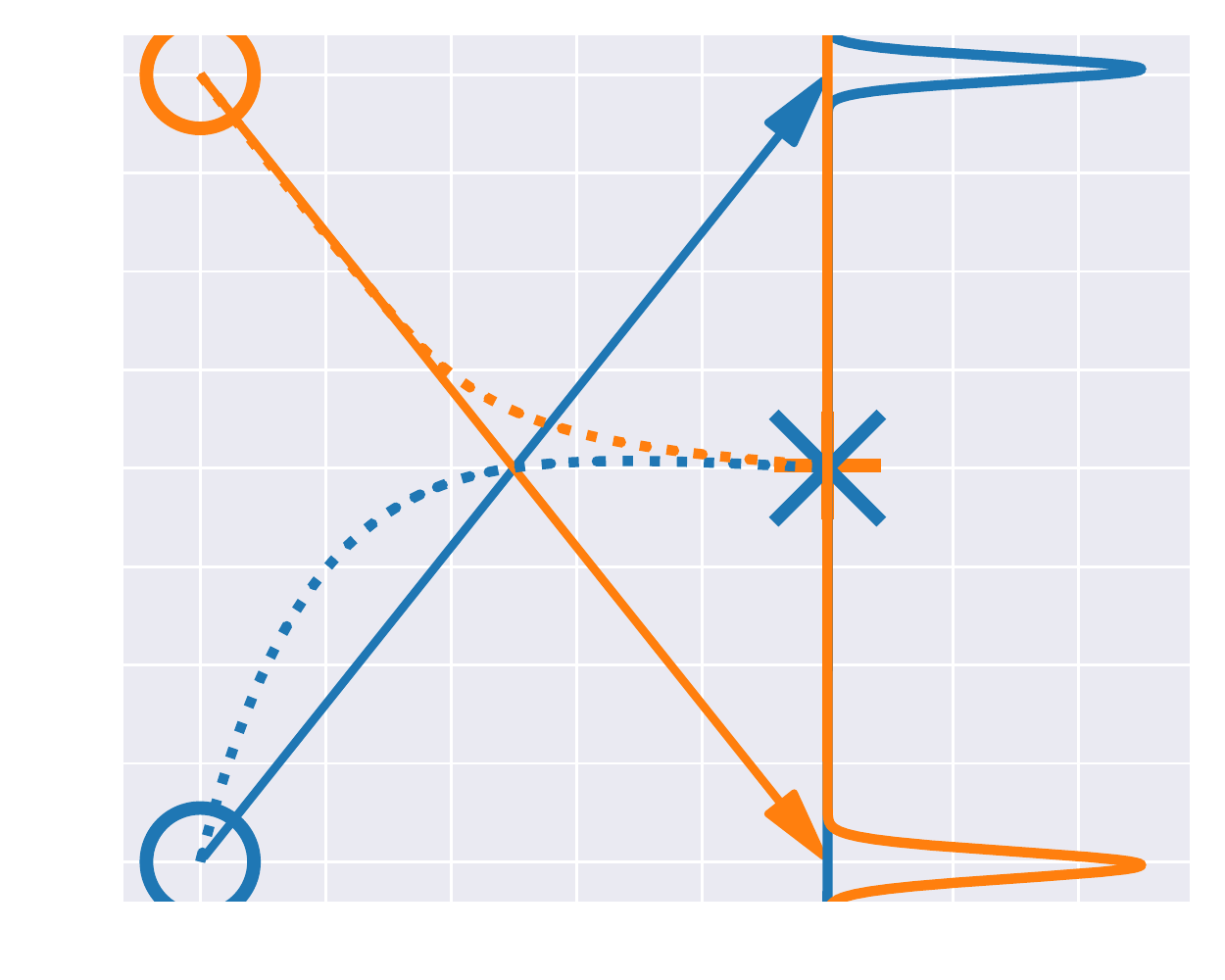}} \\
    \subfigure[Splitting]{\label{fig:fail:split:o}\includegraphics[height=22.5mm]{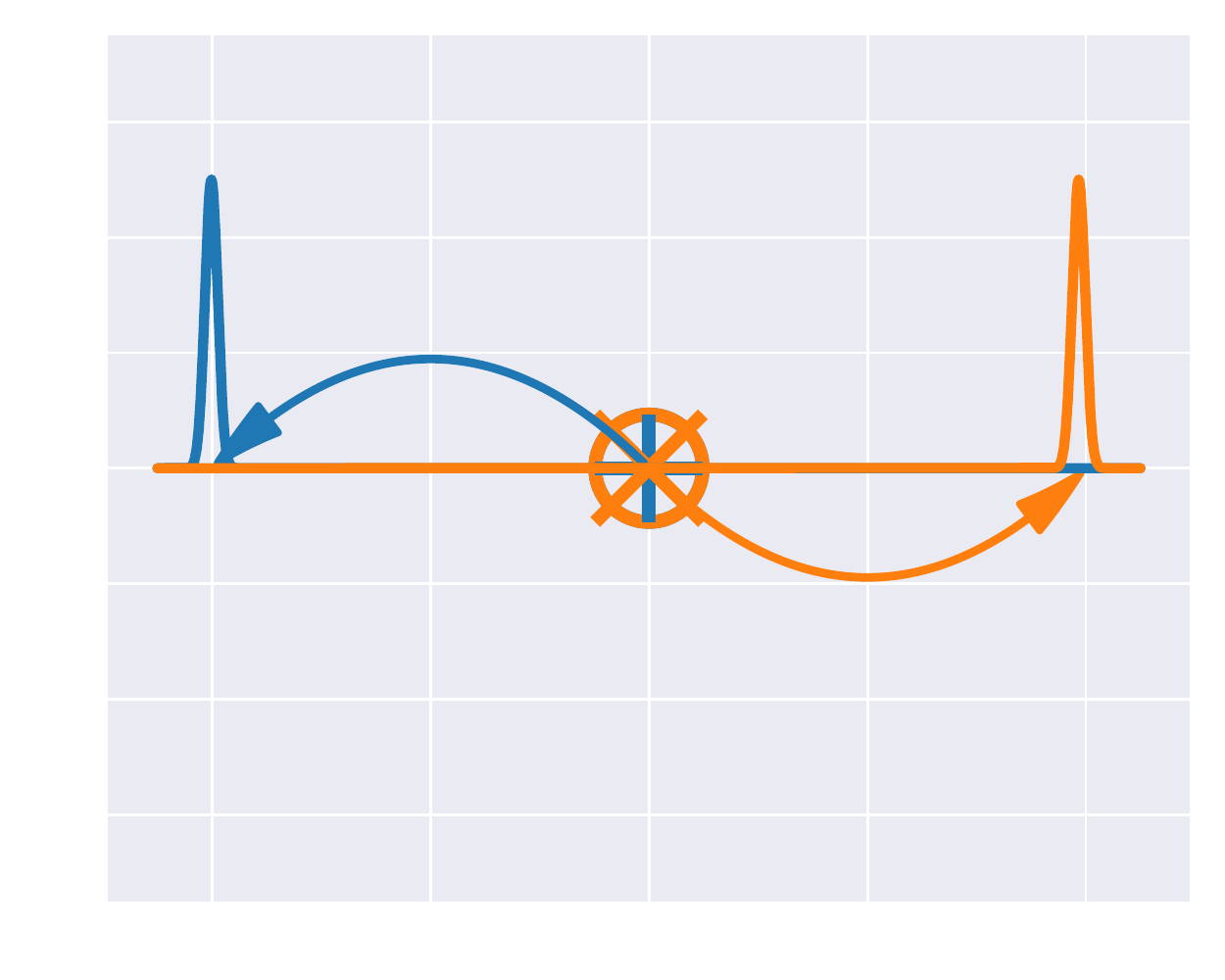}}
    \subfigure[Splitting (t)]{\label{fig:fail:split:t}\includegraphics[height=22.5mm]{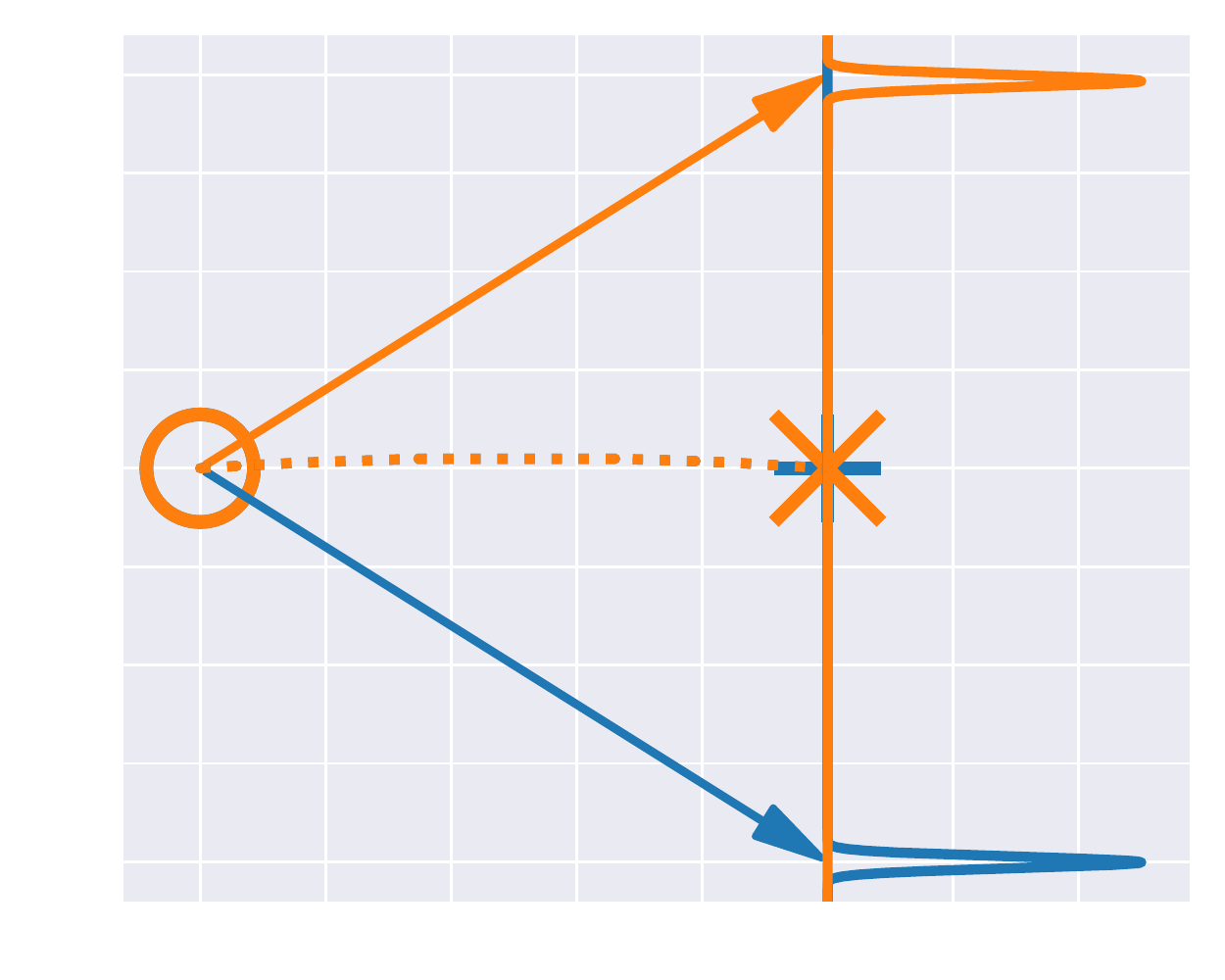}} \\
    \subfigure[Scaling]{\label{fig:fail:scale:o}\includegraphics[height=22.5mm]{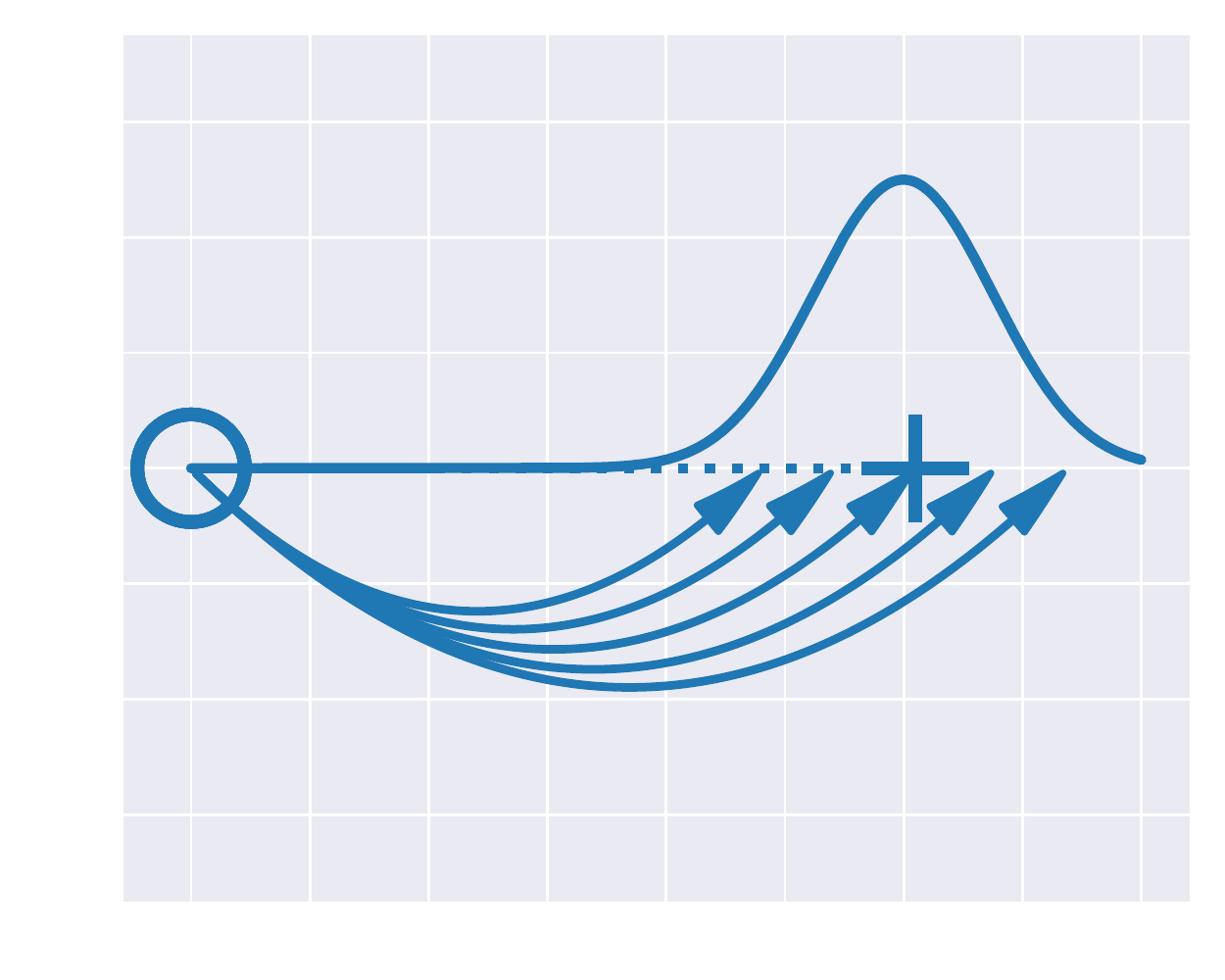}}
    \subfigure[Scaling (t)]{\label{fig:fail:scale:t}\includegraphics[height=22.5mm]{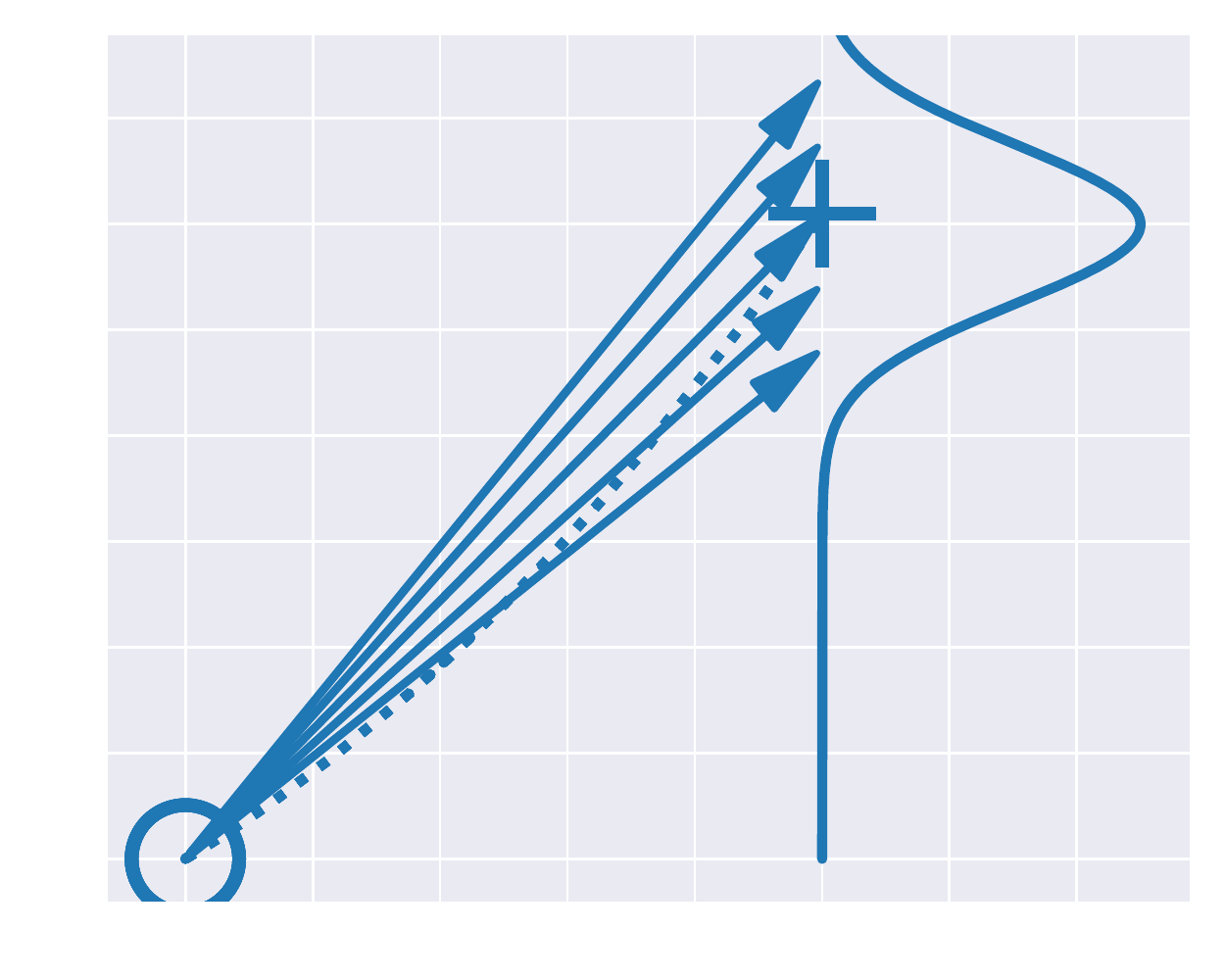}} 
    \caption{This figure illustrates failure cases for NODE models in their original space (left) and particle evolution over the integration interval (right). In these figures colour indicates class membership, the solid arrows link the start ($\circ$) and end ($\times\,+$) positions of particles (\ie ground truth data) and the dotted lines depict the learnt transportation by \ac{NODE} models. The Gaussian \acp{PDF} represent the predictive distributions of the proposed method.}
    \label{fig:fail}
    \vspace{-3em}
\end{wrapfigure}
A simple and elegant extension called \acp{ANODE} is proposed. These are experimentally shown to be a more expressive and stable model choice capable of solving the crossing problem. 

\Cref{fig:fail} illustrates \ac{NODE} failure cases. 
Crossing \cite{dupont2019augmented} tasks cannot be solved by \acp{NODE} since their advections constitute homomorphisms for which crossing is forbidden. However, \ac{ANODE} models find solutions by lifting the representation out of the original feature domain. Additionally, neither \ac{NODE} nor \ac{ANODE} models can solve splitting problems (where a single datapoint maps to more than one target) or account for uncertainty and hence they always map to a single target regardless of the variance of the system (\cf scaling cases in the figure). 

The shortcomings of baseline approaches are addressed with a probabilistic approach to \ac{ODE} problems in this work. We use indoor localisation as a motivating application which seeks to find a mapping to a relative coordinate system. We propose the use of \ac{NODE}-like models to learn typical patterns of movement and later to deploy these to forecast future behaviour. The three cases in \cref{fig:fail} are oftern encountered in behavioural modelling as follows: crossing: residents will never take the exact same path between two locations, and exemplar paths are likely to cross at least once; splitting: residents have the option to turn left, right, move forward or backtrack on their trajectory with their latent intent behind their motion determining this decision; and scaling: unobservable factors influence paths walked by residents, and even under identical starting conditions a resident may move at different speeds/directions along the same path. 

Our main contribution is the introduction and exposition of \acp{SVFM} within the \ac{NODE} paradigm, and the introduction of trajectory-focused losses that encourage simpler and less distorted solutions. Their utility is primarily enabled by the uncertainty that is posited on both the vector fields themselves and on selection of vector fields from an ensemble, but is also supported by loss functions that encourage simpler solutions to the \ac{IVP} and capture richer context in forecasting tasks. We demonstrate their capability in solving the above three fundamental tasks where baseline models fail.
Although the proposed methods are developed specifically with forecasting in mind, we also demonstrate their utility in classical classification tasks.

\section{Proposed methods}
\label{section:methods}


\subsection{Vector field mixtures} 

\begin{wrapfigure}{r}{0.5\textwidth}
    \centering
    \vspace{-7.5em}
    \subfigure[VF unit]{\label{fig:node}\includegraphics[height=33mm]{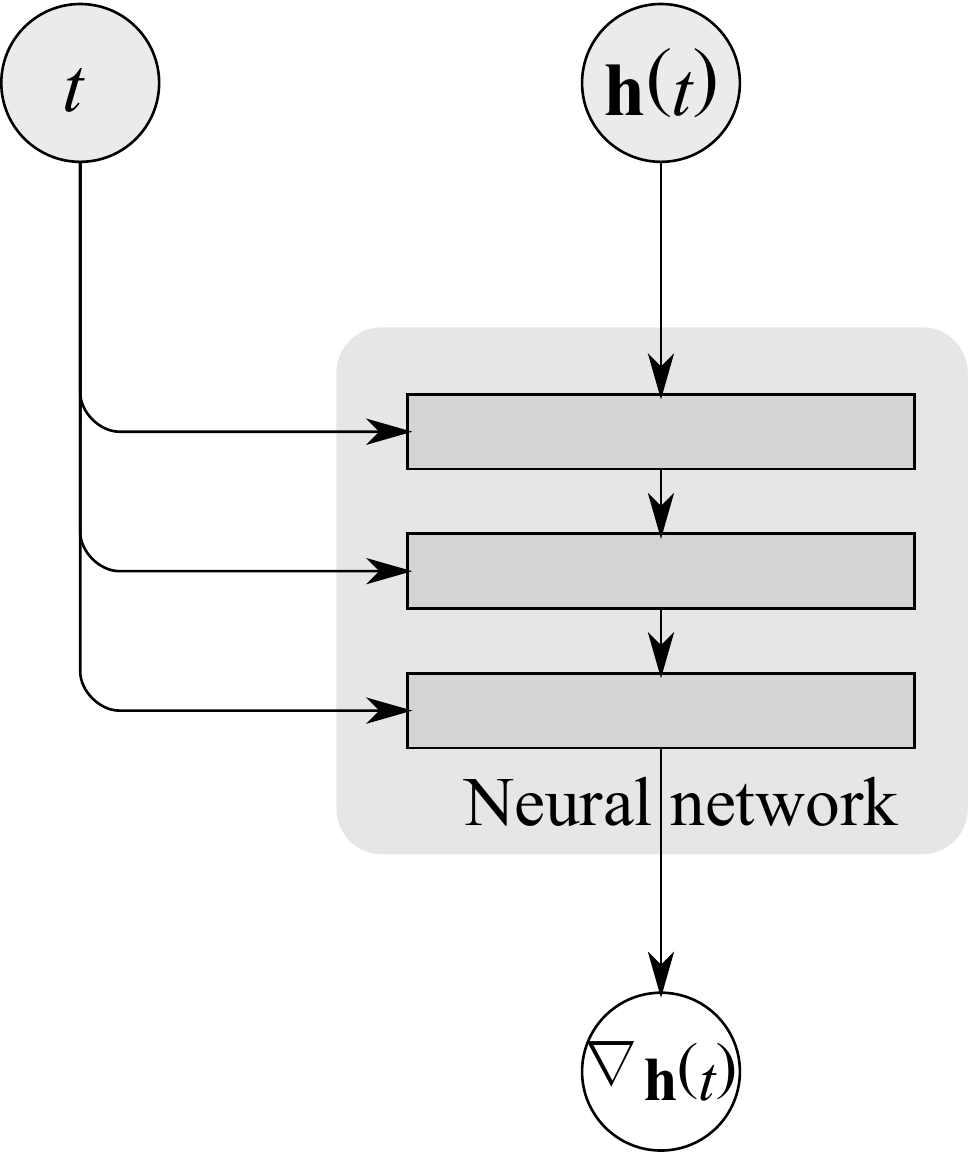}} ~
    \subfigure[VFM unit]{\label{fig:mix}\includegraphics[height=33mm]{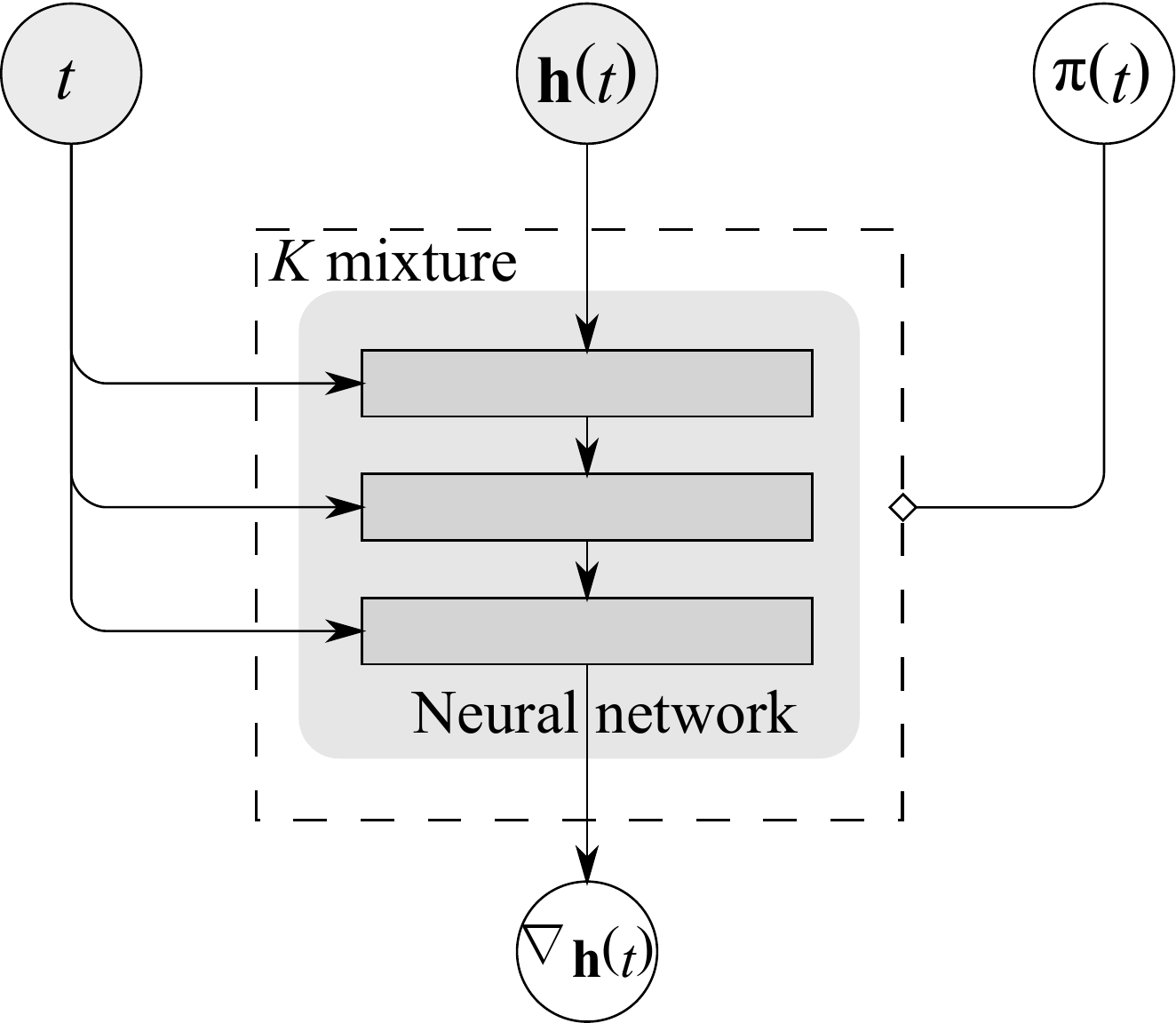}} 
    \caption{Architecture of VF and VFM units.}
    \vspace{-1em}
    \label{fig:methods}
\end{wrapfigure}
The graphical model representing the basic \ac{VF} unit is shown in \cref{fig:node}. The input variables ($t$ and $\h(t)$) are denoted by shaded nodes indicating that these values are observed, whereas the output node ($\nabla \h(t)$) is unshaded, indicating that it is latent. The assumption here is that the $i$-th \ac{VF} is conditionally independent of its history given its hidden state. In what follows we use the notation $f_g(\h(t), t; \thetab^{(g)})$ to represent a \ac{NODE} unit on an example latent variable $g$, whereas $f^{(k)}(\cdot)$ specifically defines the $k$-th \ac{VF} from the \ac{VFM}.

Using gate notation \cite{minka2009gates}, a \ac{VFM} with $K$ components is shown in \cref{fig:mix}. The new latent variable, $\pib(t) \in \triangle^K$ (where $\triangle^K$ defines the $K$-simplex) is a discrete probability distribution over component membership. The manner in which this distribution is specified is a key consideration with these models. Two approaches for defining $\pib(t)$ and propagating uncertainty over the integration interval are outlined below. 

\textbf{Pick and stick.} This operates under the assumption that component membership is constant over the forecast horizon, \ie $\pib(t_i) = \pib(t_{*}) ~~ \forall ~i ~ (0 \leq i \leq T)$. Several choices for specifying $t_{*}$ may be considered, but if mixture dynamics are well specified at $t_0$ we can set

\begin{align}
    \pib(t_{*}) = \pib(t_0) = f_{\pib_{t_0}}(\h(t_0), t_0; \thetab^{(\pib_{t_0})})
\end{align}


%
%
%
%

\textbf{Forward filtering.} Forward filtering is a technique used in dynamic systems to estimate belief based on a history of evidence, and it is often used in dynamic models like \acp{HMM} \cite{rabiner1986introduction} and \acp{CRF} \cite{sutton2012introduction}. It is incorporated into our model directly by modeling transition and emission dynamics of component membership as follows

\begin{align}
    \label{eq:transition}
    \Psib(t_i) &= f_{\Psib}(\h(t_i), t_i; \thetab^{(\Psib)})\\
    \label{eq:emission}
    \psib(t_i) &= f_{\psib}(\h(t_i), t_i; \thetab^{(\psib)})
\end{align}

where each of the $K$ rows of $\Psib(t_i)$ and $\psib(t_i) \in \triangle^K$. Forward filtering aggregates belief as follows

\begin{align}
    \pib(t_i) \propto \Psib(t_i)^\top\left( \psib\left(t_i\right) \odot \pib\left(t_{i-1}\right) \right)
\end{align}

\noindent where $\odot$ depicts the Hadamard product, $\top$ the matrix transpose, and equality is achieved with normalisation. 
A complementary interpretation of forward filtering is that of a \ac{RNN} \cite{zheng2015conditional} that grows probabilistically through the forecasting horizon.

\subsection{Stochastic vector fields}

\begin{wrapfigure}{r}{0.2\textwidth}
    \centering
    \vspace{-5em}
    \includegraphics[height=33mm]{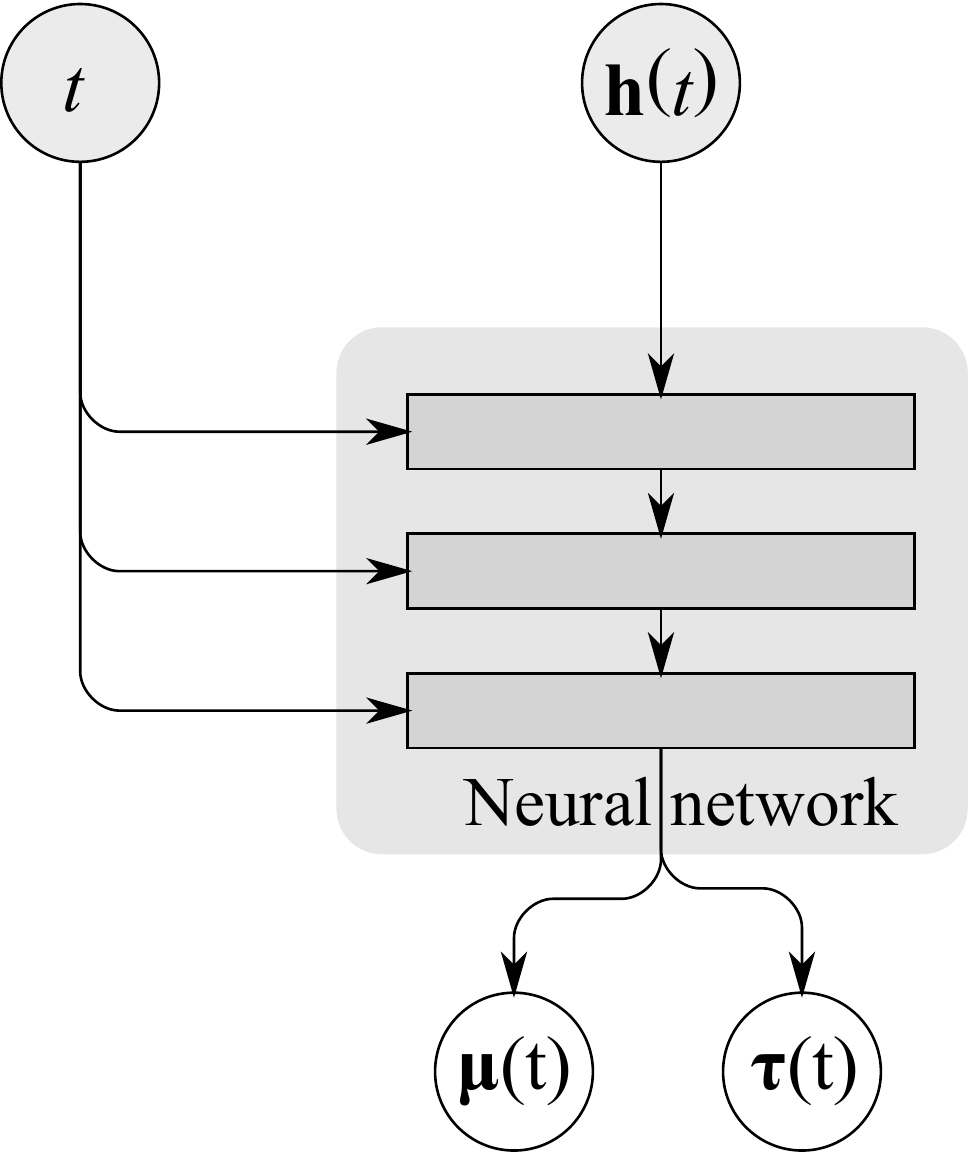}
    \caption{SVF unit.}
    \vspace{-2em}
    \label{fig:uvf}
\end{wrapfigure}
Uncertainty is introduced to \acp{VF} with the mean ($\mub(t)=g_{\mub}( f_{\z}(t), t; \thetab^{(\mub)} )$) and variance ($\taub(t)=g_{\taub}( f_{\z}(t), t; \thetab^{(\taub)} )$) generating functions that both arise from a common representation $f_{\z}(t)=g_{z}( \h(t), t; \thetab^{(z)} )$, see \cref{fig:uvf}. It will be convenient to decompose \acp{VF} into the length ($v(t)$) and orientation ($\u(t)$) of the \ac{VF} and to impose uncertainty on these separately:

\begin{align}
    \label{eq:dir_uncertainty}
    \u(t) &\sim \mathcal{N}_s\left( \mub^{(\u)}(t), \taub^{(\u)}(t) \right) \\
    \label{eq:len_uncertainty}
    \log v(t) &\sim \mathcal{N}\left( \log \mub^{(v)}(t), \taub^{(v)}(t) \right) 
\end{align}

where $\mub^{(\u)}(\cdot)$, $\taub^{(\u)}(\cdot)$, $\mub^{(v)}(\cdot)$ and $\taub^{(v)}(\cdot)$ are the mean and variance representations of the direction and length variables respectively, $\mathcal{N}(\cdot)$ is a Gaussian distribution and $\mathcal{N}_s(\cdot)$ is a distribution defined on a $D$-sphere. Samples from both distributions can be transformed into \acp{VF} with the inverse transformation to that used above, $\nabla \h(t) = \exp\{v(t)\} \u(t)$. 

In principle one may wish to constrain the variance of \ac{SVF} to ensure that $\u(t) \cdot \mub^{(\u)}(t) > 0$ since, with large variance, samples from the directional distribution (\cref{eq:dir_uncertainty}) will be reversed with respect to the expected direction of the \ac{VF}. We believe this directional preservation property is important since it preserves the expected inductive bias of the \ac{SVF} for each sample, which tends to give lower variance samples. A similar motivation led to us imposing uncertainty on length in the log domain. Although noise can be imposed on \acp{VF} without the decomposition (\eg see \citet{otto2010uncertain}), the proposed decomposition naturally orientates the distribution along the expected direction of the \ac{VF}. 

\subsection{Loss functions}

Two factors directly affect the \ac{NFE} when solving \acp{IVP}: the total distance moved by the particle and the `straightness' of its trajectory. In this section we motivate several losses, and in particular transportation and variance losses have a minimising effect \acp{NFE}. 

\textbf{\ac{MDL}.} \ac{SVFM} models produce mixture distribution as outputs, and these are reconciled into a mixture density loss as follows

\begin{align}
    \label{eq:mdn}
    \mathcal{L}_{\text{MDLoss}} = -\log \sum_{k=1}^K p\left(\h\left(t_i\right) | \mub^{(k)}\left(t_i\right), \taub^{(k)}\left(t_i\right) \right) \pib^{(k)}\left(t_i\right)
\end{align}

\noindent where $p$ is the density of the output distribution. This loss resembles those given in Gaussian mixture models and mixture density networks \cite{bishop1994mixture}.

\textbf{\ac{TL}.} A regulariser based on path length is defined as follows

\begin{align}
    \label{eq:tl}
    \mathcal{L}_{\text{TLoss}} = \frac{1}{T} \sum_{i=1}^T\| \h(t_{i}) - \h(t_{i-1}) \|_2^2
\end{align}

The minimal possible $\mathcal{L}_{\text{Dist}}$ between $\h(t_0)$ and $\h(t_T)$ is  $\| \h(t_{T}) - \h(t_{0}) \|^2_2$ and this is achieved when all \ac{VF} evaluations point in the same direction, \ie $\nabla \h(t_i) = \nabla \h(t_0) s_i ~ \forall s_i \in \mathbb{R}_{+}$.

\textbf{\acf{DV}.} \acp{VF} can be encouraged to point in the same direction by regularising their variance. Letting $\mathbb{E}[\nabla \h(t)] = \frac{1}{T}\sum_{i=1}^T\nabla \h(t_i)$ be the expected direction, \ac{VF} variance loss is

\begin{align}
    \label{eq:dp}
    \displaystyle\mathcal{L}_{\text{VLoss}} = \frac{1}{T}\sum_{t=1}^T \left\| \nabla\h(t_i) - \mathbb{E}\left[\nabla \h(t)\right] \right\|^2_2 
\end{align}

Combined transportation distance and directional variance losses are termed \ac{PLL}.

\textbf{\ac{APL}.} In forecasting problems, a datapoint is a sequence of $T'$ measurements and timestamps. We define $\X\in\mathbb{R}^{T'\times D}$ to be the matrix of measurements and $\mathbf{t}_\X \in \mathbb{R}^{T'}$ to be the corresponding timestamps. The total path loss is then defined as follows

\begin{align}
    \label{eq:apl}
    \mathcal{L}_{\text{FLoss}} =  \frac{1}{T}\sum_{i=1}^T \ell\left( \h(t_i), \interp\left( \X, \mathbf{t}_\X, t_i \right) \right)
\end{align}

\noindent where $\interp$ is an interpolation function that estimates data at $t_i$, and $\ell$ is a secondary loss that penalises $\h(t_i)$ that are far from the interpolated target. We choose cubic interpolation in our experiments and the choice of $\ell$ dependent on the model; mean squared error is used with baseline methods and \ac{MDL} is used with the proposed model. 

%


\section{Sampling and SVFM relationships}

\begin{figure}[t]
    \centering
    \subfigure[Mixture uncertainty]  {\label{fig:ex:mix}\includegraphics[height=32.5mm]{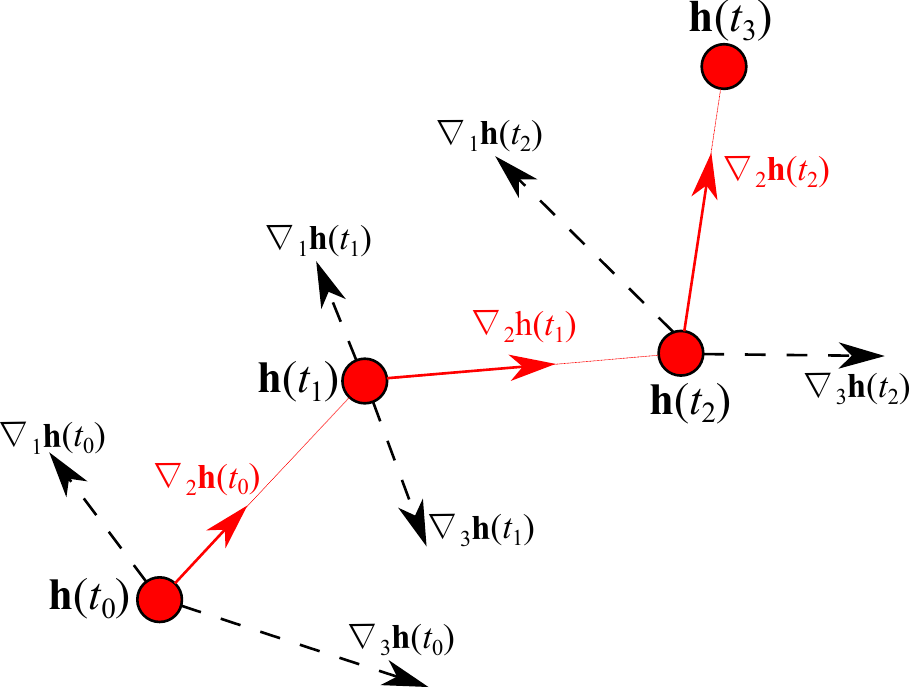}} ~
    \subfigure[Length uncertainty]   {\label{fig:ex:len}\includegraphics[height=32.5mm]{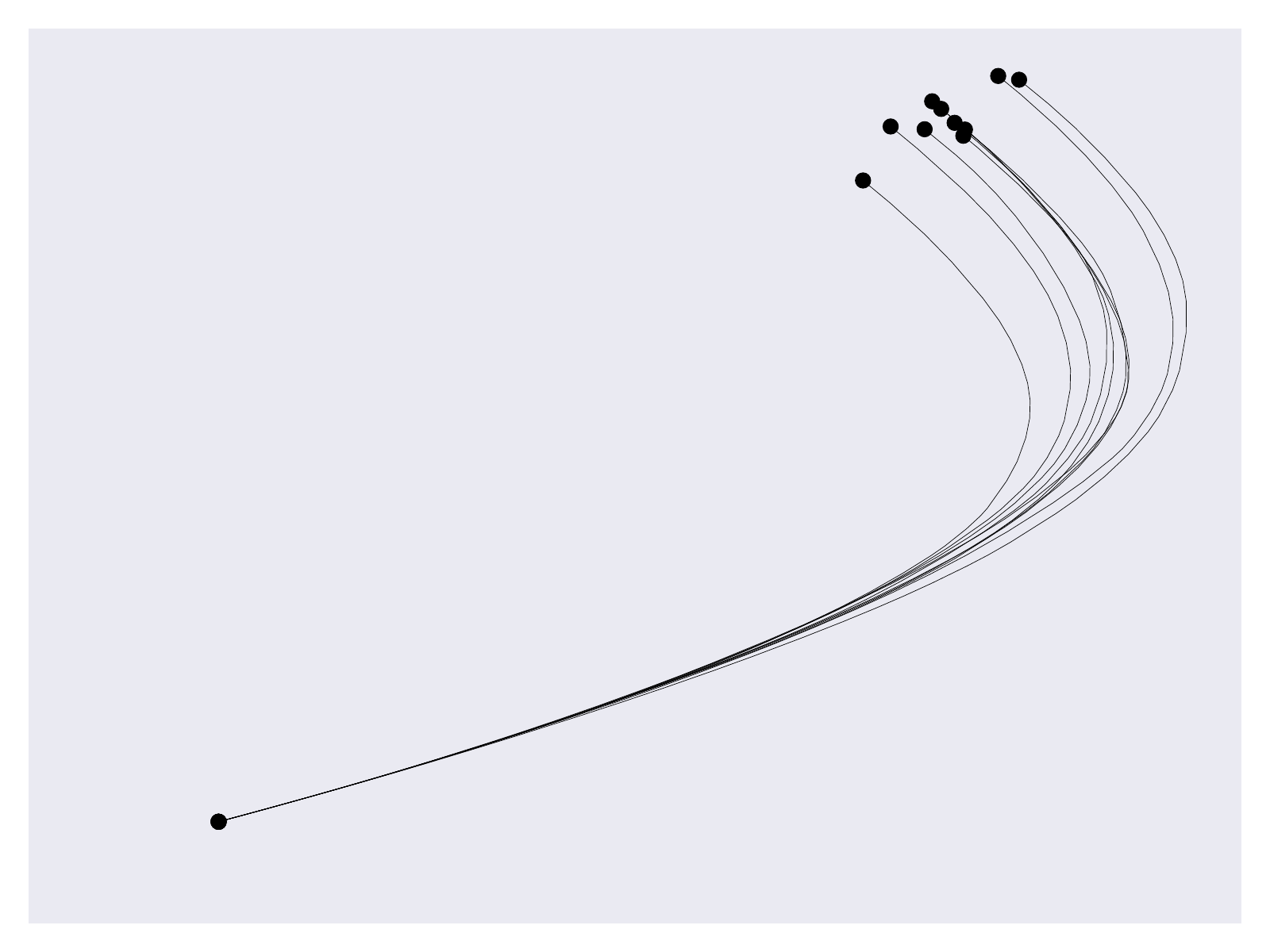}} ~
    \subfigure[Direction uncertainty]{\label{fig:ex:dir}\includegraphics[height=32.5mm]{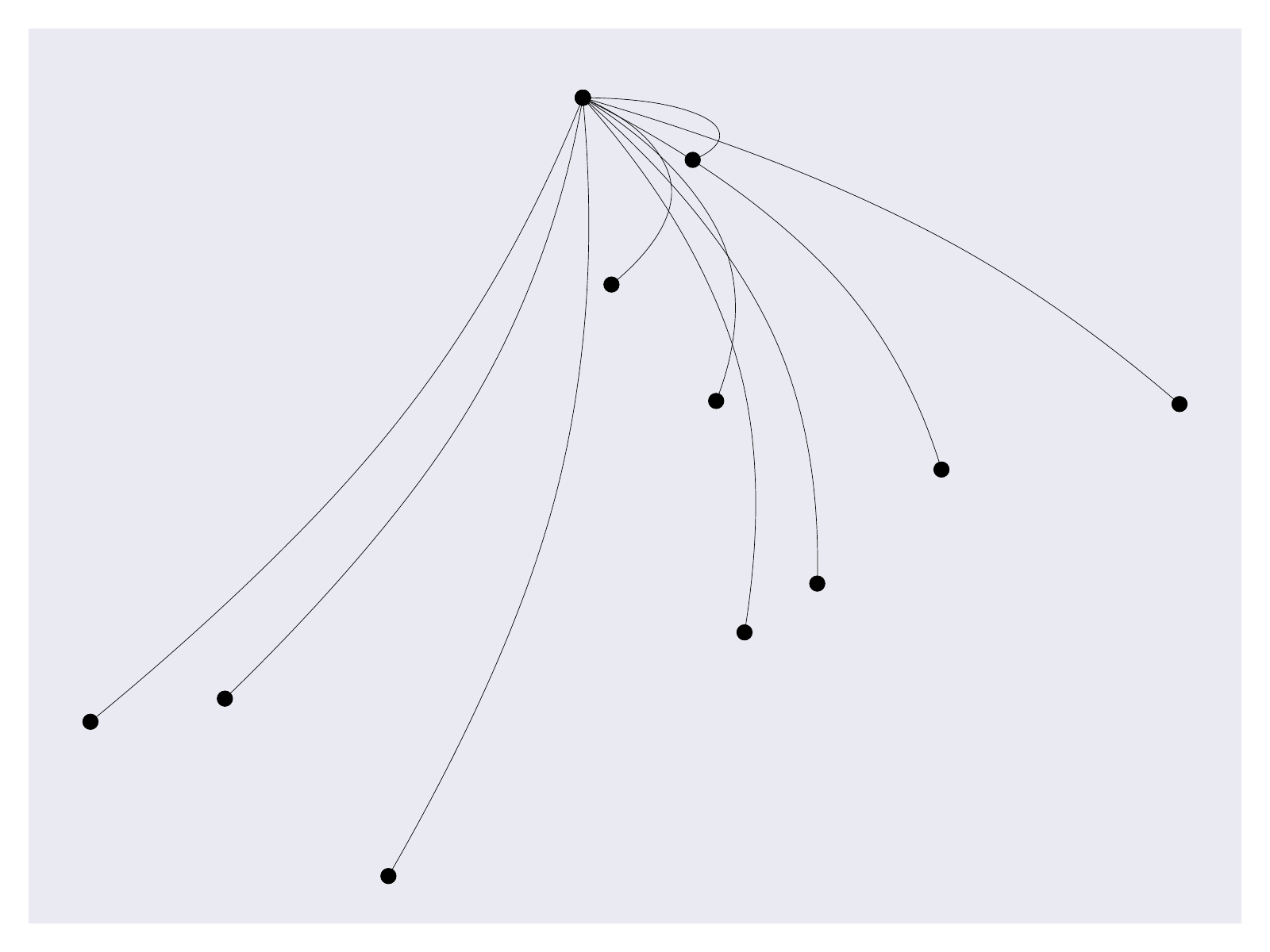}} 
    \caption{Illustration of three forms of uncertainty on randomly initialised VF units.}
    \label{fig:uncertainty}
\end{figure}

\begin{wrapfigure}{r}{0.3\textwidth}
    \centering
    \vspace{-2em}
    \includegraphics[height=60mm]{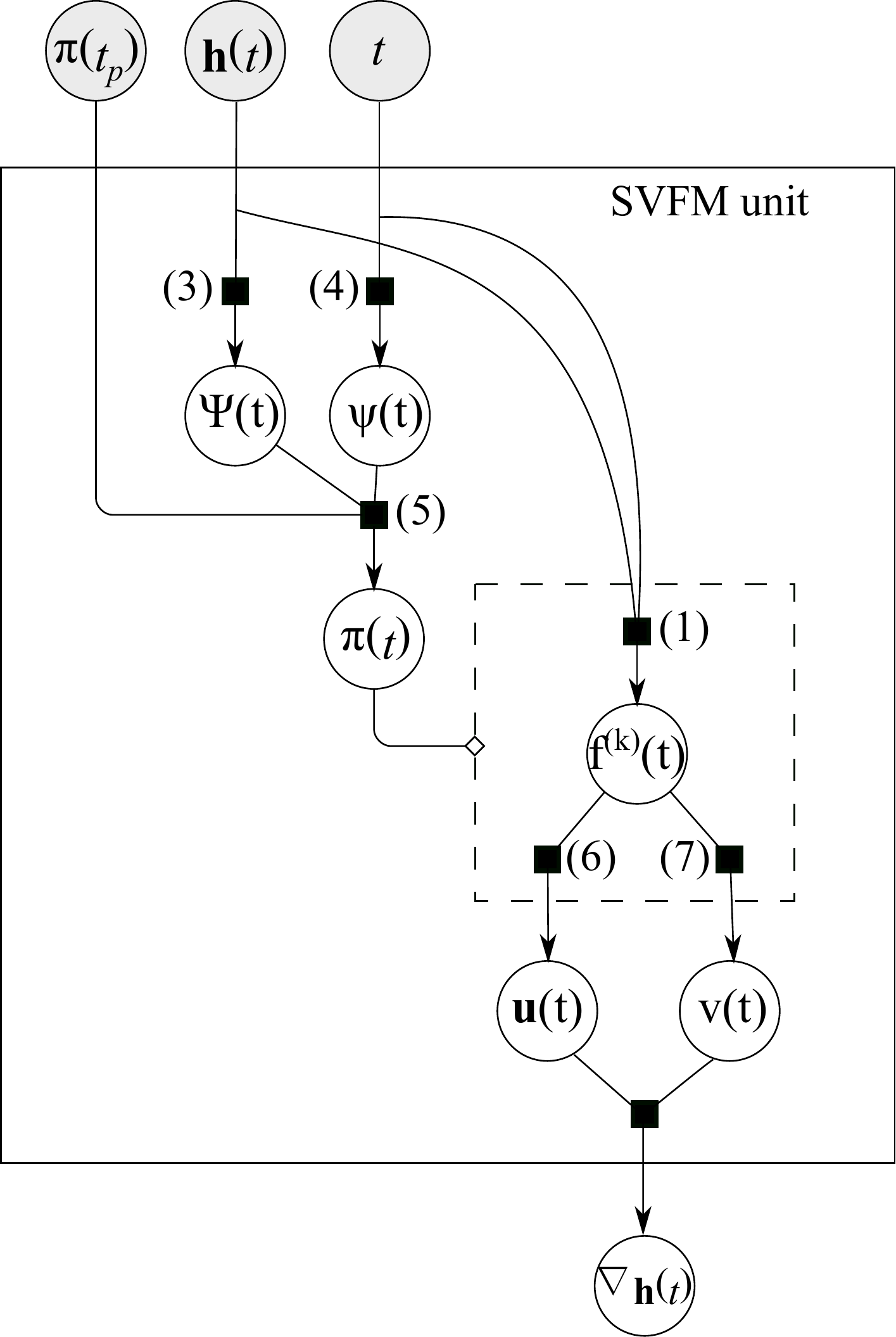}
    \caption{SVFM unit.}
    \vspace{-1em}
    \label{fig:umnode_unit}
\end{wrapfigure}

%
\Cref{fig:ex:mix} illustrates branching patterns that are characteristic of \ac{VFM} solutions. The initial state $\h(t_0)$ is shown by the red node on the bottom left, and its trajectory to $\h(t_3)$ is highlighted in red. At each evaluation position one of three \acp{VF} is dynamically sampled from the distribution $\pib(t_i)$. Although we only consider myopic approaches here, \ie steps are taken based on samples from $\pib(t_i)$, other approaches (including particle filtering and reinforcement learning) will be explored in future work. Exhaustive search is prohibitive since it is exponential on the number of evaluation points. Ten samples from uncertain \acp{VF} are shown in \cref{fig:ex:dir,fig:ex:len}. Scaled \acp{VF} (\cref{fig:ex:len}) which always initially follow the same trajectory will tend to `fan out' as curvature is encountered whereas directional uncertainty (\cref{fig:ex:dir}) tends to fan out immediately. The full factor graph of the proposed approach is shown in \cref{fig:umnode_unit}. \ac{NODE} units are depicted by the black squares in this figure.

%
The proposed model, \ac{SVFM}, encapsulates all of the properties shown in \cref{fig:umnode_unit}. Sequences of samples from \acp{SVFM} are likely to vary significantly even under the same contextual conditions, and this will cause problems for \ac{ODE} solvers. Since the classic \ac{RK4} method requires four function evaluations in determining the next direction \cite{kutta1901beitrag} and the \ac{DP} method requires six \cite{dormand1980family} their convergence guarantees do not hold since \acp{SVFM} are discontinuous everywhere \cite{sarkka2019applied}, but with careful scheduling it is possible for these guarantees to hold at each step. Our approach is to record the random state at the beginning of the evaluation campaign, and to reset the random number generator's state to the recorded state whenever samples need to be drawn during the campaign with \cite{ruiz2016generalized,jang2016categorical}. Experimentally, this trick significantly reduces the number of function evaluations required with \ac{DP} solvers that are used in our experiments.

\begin{wrapfigure}{r}{0.5\textwidth}
    \centering
    \vspace{-1em}
    \includegraphics[height=30mm]{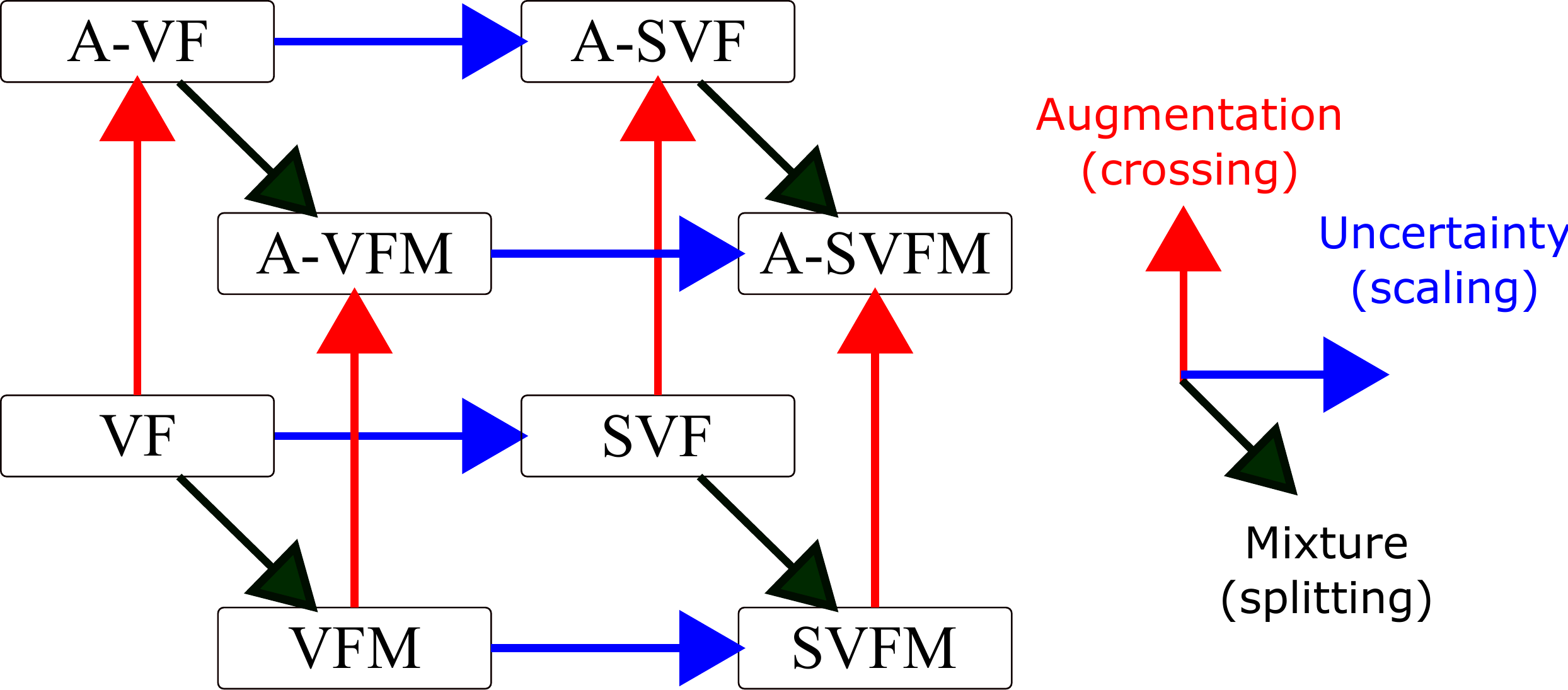}
    \caption{Relationships between discussed models.}
    \vspace{-2em}
    \label{fig:cube}
\end{wrapfigure}

The incorporation of uncertainty and mixture distributions to \acp{VF} are the key factors that help solve scaling and splitting cases and also facilitate behavioural modelling. \Cref{fig:cube} highlights the relationship between stochastic (S-, blue), mixture (-M, black) and augmented (A-, red) \cite{dupont2019augmented}  models. This figure not only shows that these three modelling choices can be used together, but that any combination can be selected. 
%

Although the most complete model is A-SVFM as it incorporates all three components, our experimentation will primarily be evaluated with \ac{SVFM} models because our interest is in forecasting and augmentation lifts the representation out of the original feature space.

We study some properties of these model classes theoretically. First we prove that \ac{NODE} and \ac{ANODE} models are unable to model splitting and scaling problems. We belive this further justifies our work since these problems are basic and (conceptually) simple components. Secondly we characterise the nature of \acp{VF} that require minimal \acp{NFE}. Although it turns out that the resulting \acp{VF} are trivial and are not useful in general, the analysis provides justification for the variance and transportation losses mentioned in the previous section. Details of these analyses can be found in the supplementary material.





\section{Results and discussion}
\label{section:results}

\begin{wrapfigure}{r}{0.375\textwidth}
    \centering
    \vspace{-2em}
    \subfigure{\includegraphics[width=\linewidth]{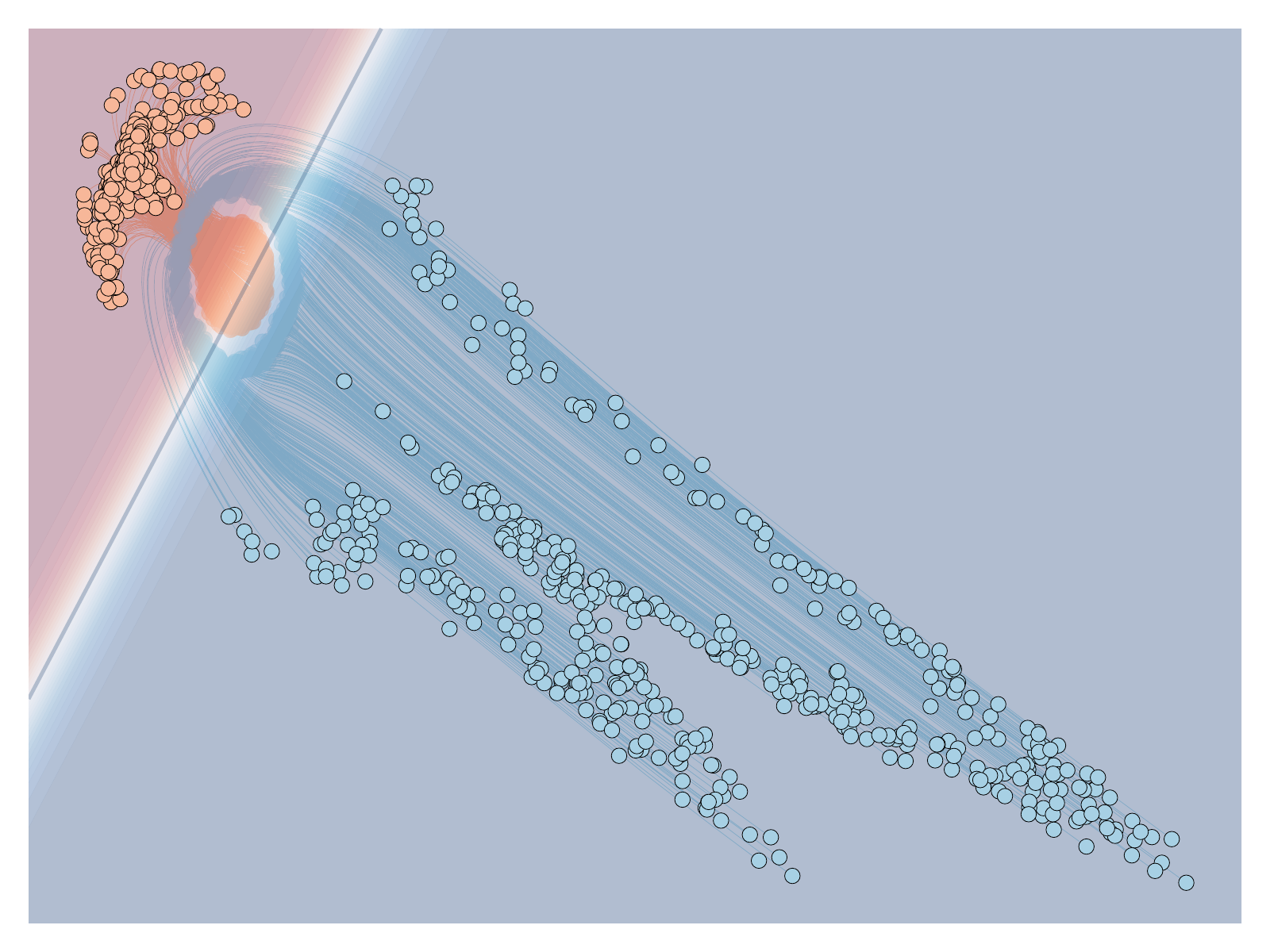}} \\ 
    \subfigure{\includegraphics[width=\linewidth]{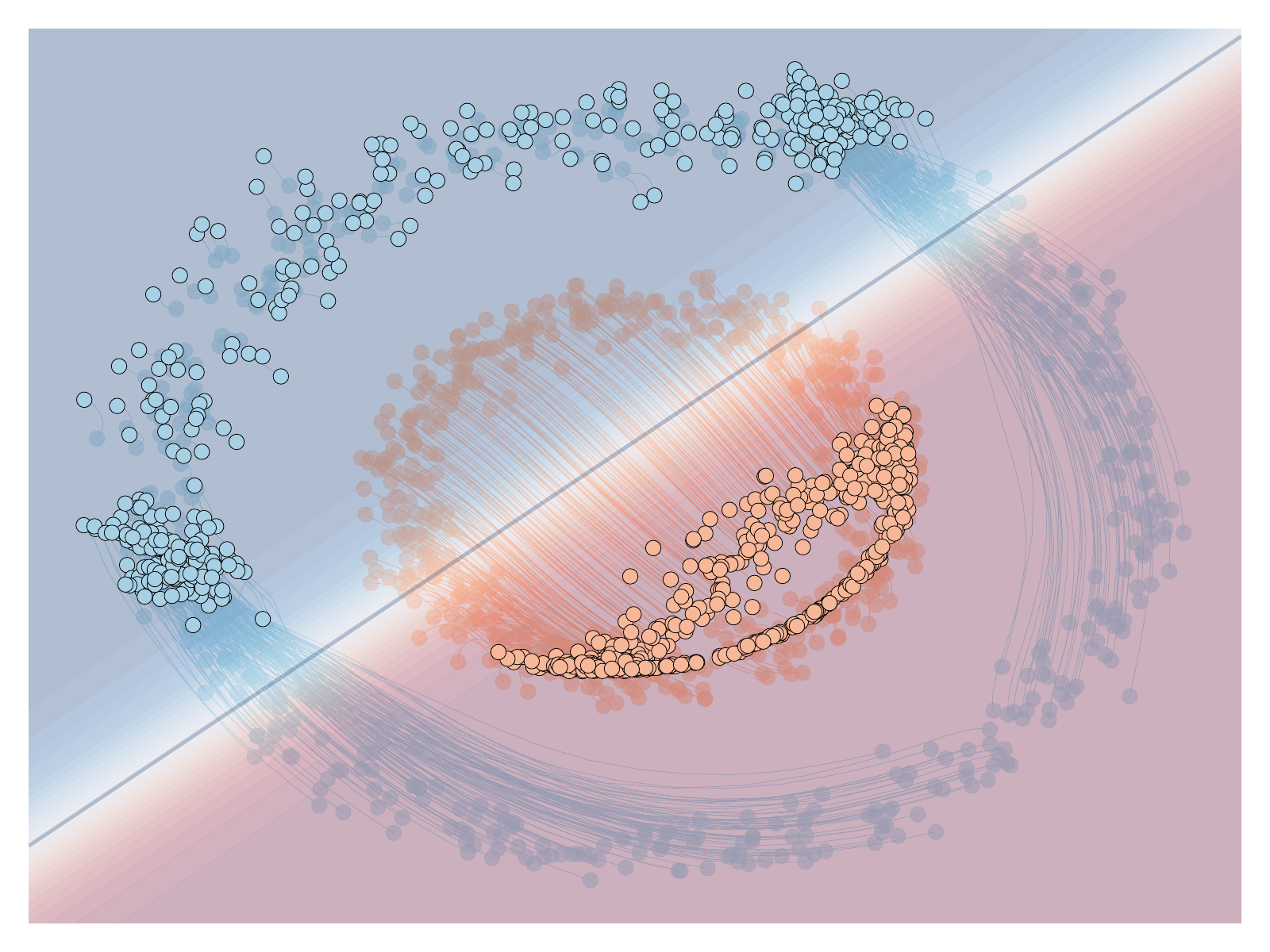}} \\ 
    \subfigure{\includegraphics[width=\linewidth]{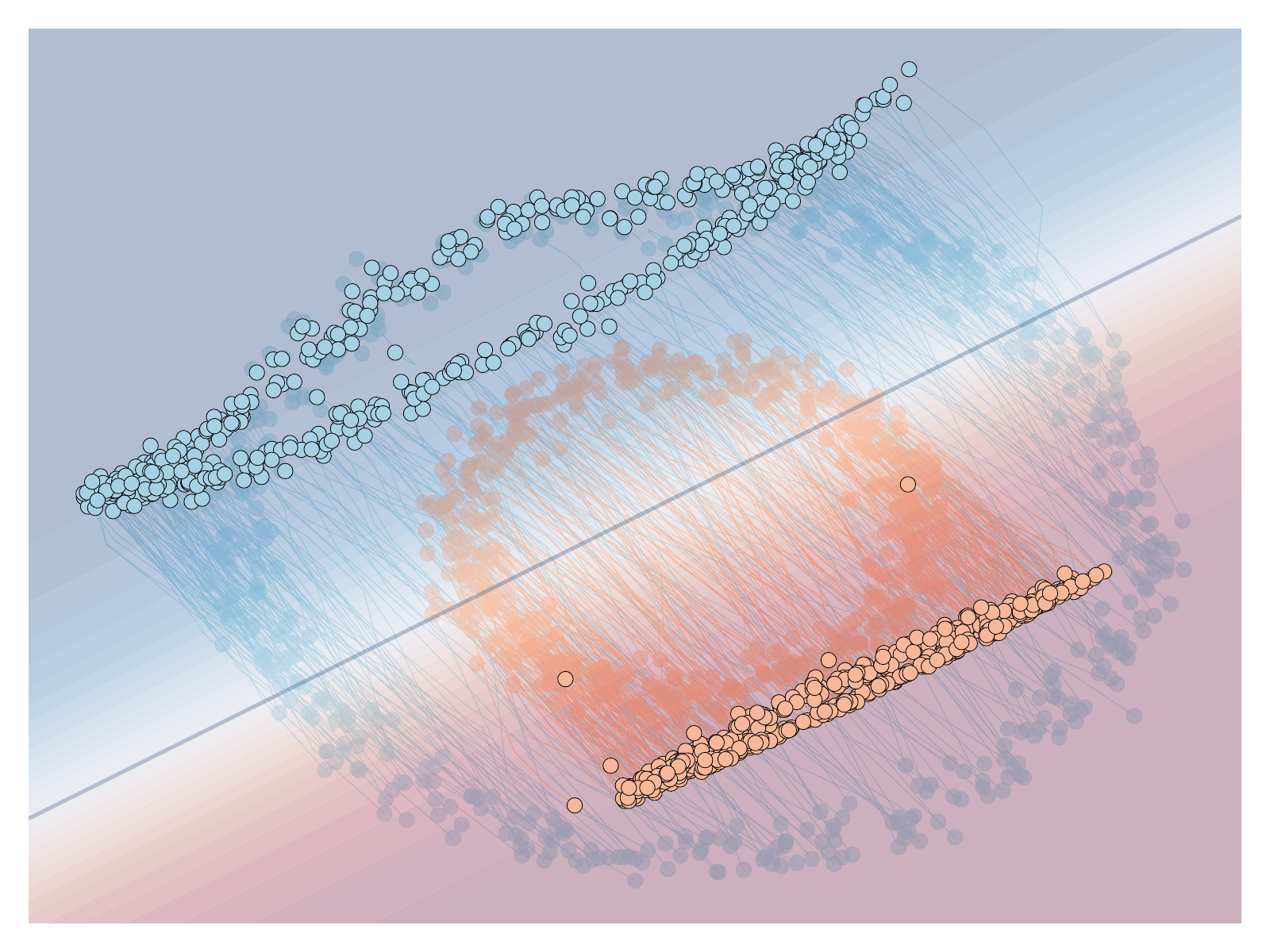}} 
    \caption{Nested circle predictions.}
    \label{fig:nested}
    \subfigure{\includegraphics[width=\linewidth]{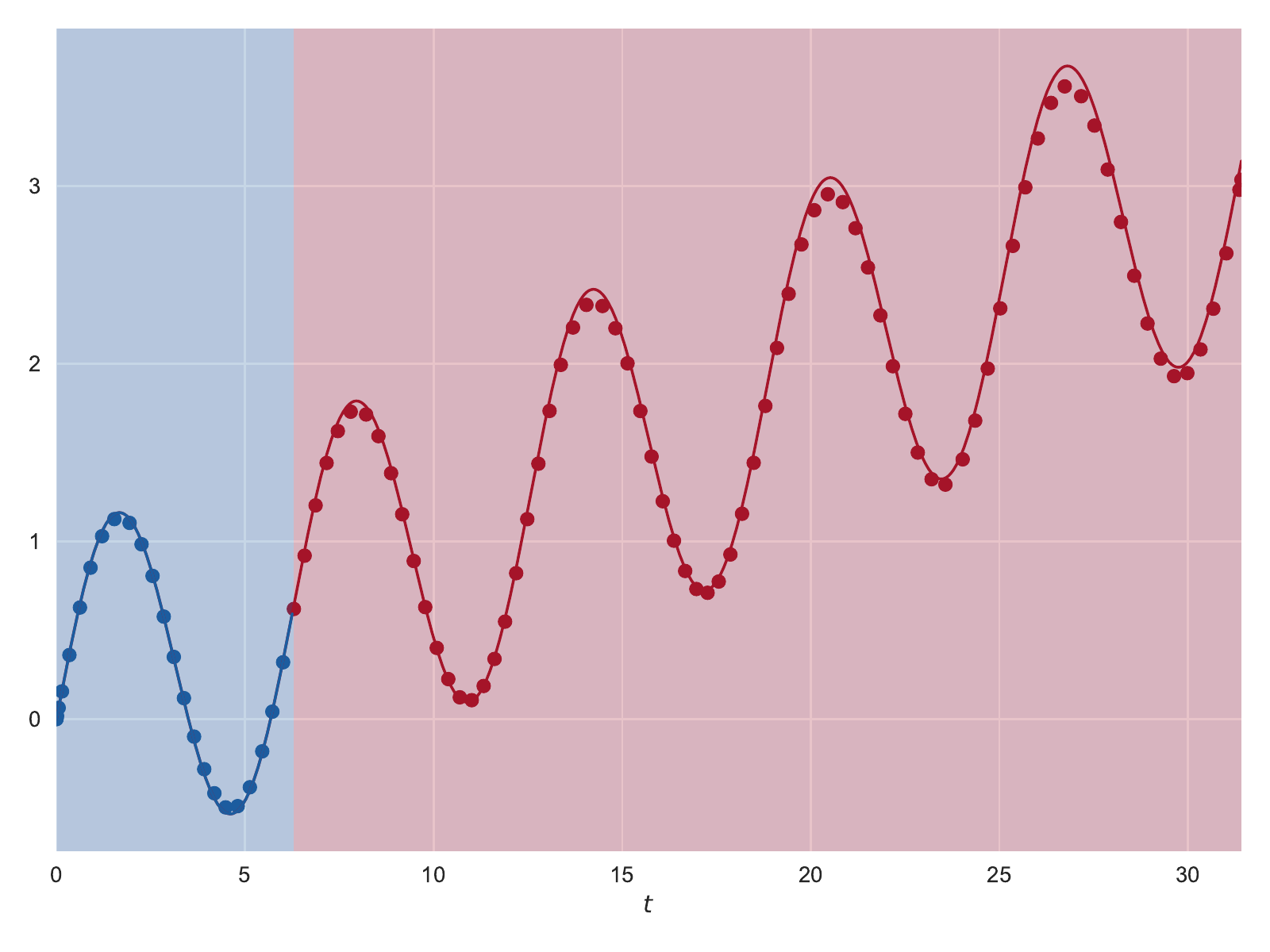}} %
    \caption{Cyclic continuation.}
    \label{fig:periodic}
    \vspace{-2em}
\end{wrapfigure}

%
\textbf{Illustrative examples.} \Cref{fig:nested} depicts transformation of \ac{VF} (top), \ac{VF} with \ac{PLL} (middle; herein called \ac{VF}-\ac{PLL}) and \ac{SVFM} (bottom) models on the nested circles dataset. The \ac{NODE} solution has learnt a complex trajectory that requires the inner circle to `break' through the outer circle. We note that particles are transported large distances through many \ac{NFE} to break through, and this jeopardises generalisation accuracy. We can reduce \ac{NFE} by incorporating \ac{PLL}. Their effect is visibly seen on the top left of the middle figure where several blue points remain stationary at their original positions and all red trajectories are close to linear. However, in solving this problem many blue datapoints have been transported \textit{around} the outer circle and these nonlinear trajectories require many \acp{NFE}. \ac{SVFM} (bottom) can be seen to produce linear transformations since the trajectory of any two particles are independent if assigned to different \ac{VF} components.

%
%
Although the example in \cref{fig:nested} (bottom) clearly demonstrates that \ac{SVFM} models facilitate linear trajectories that need fewer function evaluations, the learnt model in this particular case has discovered a trivial solution since $\pib(t_0)$ is directly associated to the class label. In the supplement we illustrate a more comprehensive example on the \ac{XOR} dataset. \ac{SVFM} cannot learn a trivial solution of this dataset but will still produces solutions requiring fewer \acp{NFE} than baselines. However, it is more challenging to visually assess linear preservation due to the nature of the \ac{XOR} dataset.

The integration variable $t$ and the interval over which it is integrated are abstract quantities that directly affect the depth of the network. In our application, we interpret $t$ as the \ac{TOD} and 
instead of parameterising our \ac{VF} networks with the time variable $t$ we use a time tuple $\mathbf{t} = (\cos(t / l), \sin(t/l))$ ($l = \frac{\pi}{12}$ if $t$ is given in hours) since it has smooth and periodic characteristics for increasing $t$. 

We call the augmentation of $t$ to $\mathbf{t}$ cyclic \ac{VF} continuation and an example of it is illustrated in \cref{fig:periodic}. Here, the function $g(t) = \sin(t) + t / 10$ (shown in the solid trace) is learnt by a \ac{NODE} mode with data from the blue interval ($0 \leq t \leq 2\pi$). The model is then tested by querying the endpoint at $t^{*}=10\pi$, a point well outside the training domain. When determining the endpoint, the \ac{NODE} model performs several function evaluations shown with the blue and red dots. We see that the periodic traits of the function have been extrapolated well beyond the training interval into the red region. Of particular note is that all intermediate datapoint evaluations are faithful to the true function which is directly encouraged by the forecasting loss. In the context of behavioural modelling, this simple example demonstrates that cyclic \ac{VF} continuation 
should allow for periodic behavioural traits to be representatively modelled over long forecast horizons. Although the demonstration here is rather straightforward, richer notions of meaningful periodic dependency will be captured from real datasets and tasks. Traditional \ac{VF} parameterisations cannot learn even this basic extrapolation task.

\begin{figure}[h]
    \centering
    \subfigure[Moons]{\label{fig:nfe:moons}\includegraphics[width=0.3\linewidth]{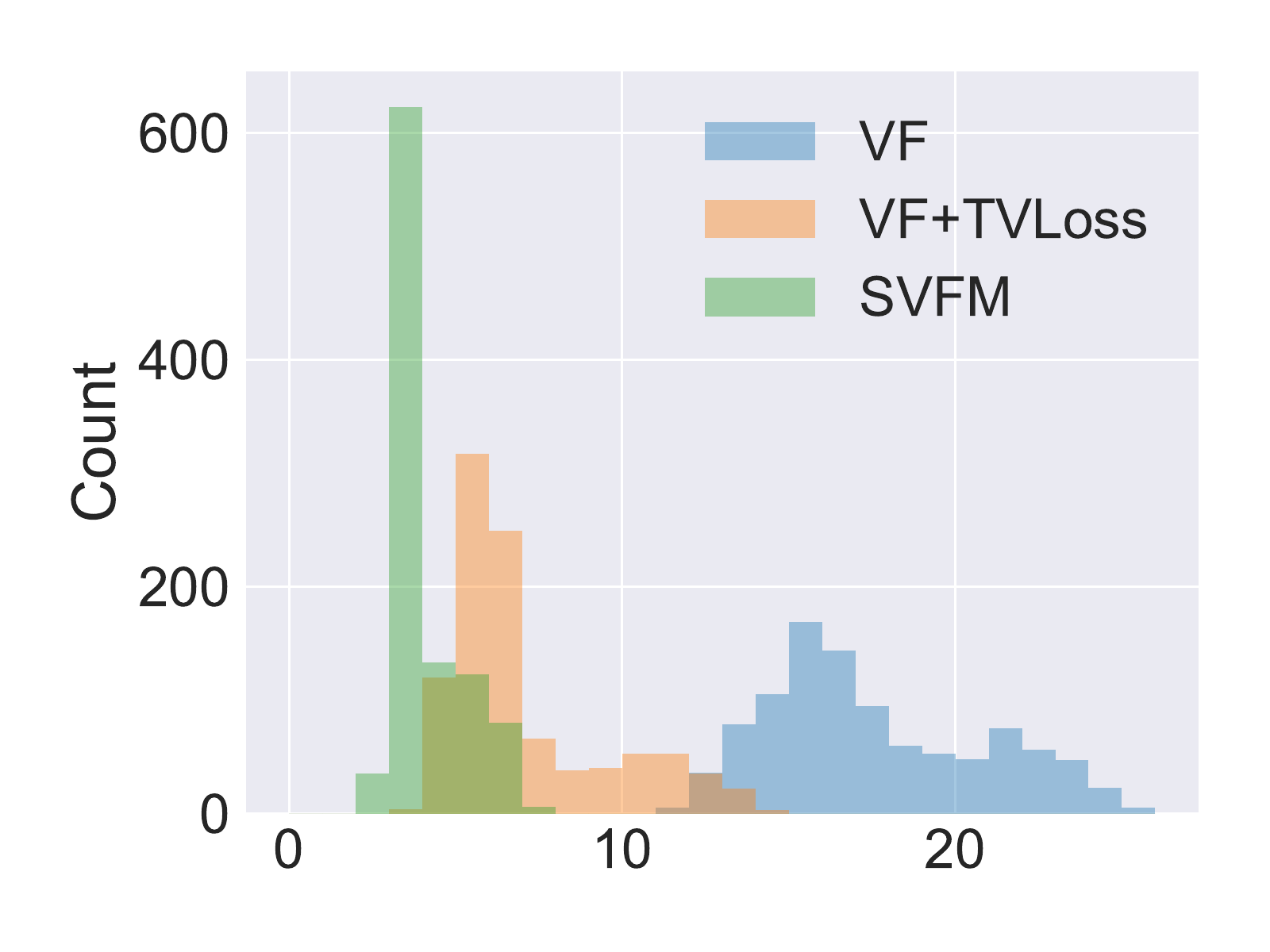}} 
    \subfigure[Circles]{\label{fig:nfe:nest} \includegraphics[width=0.3\linewidth]{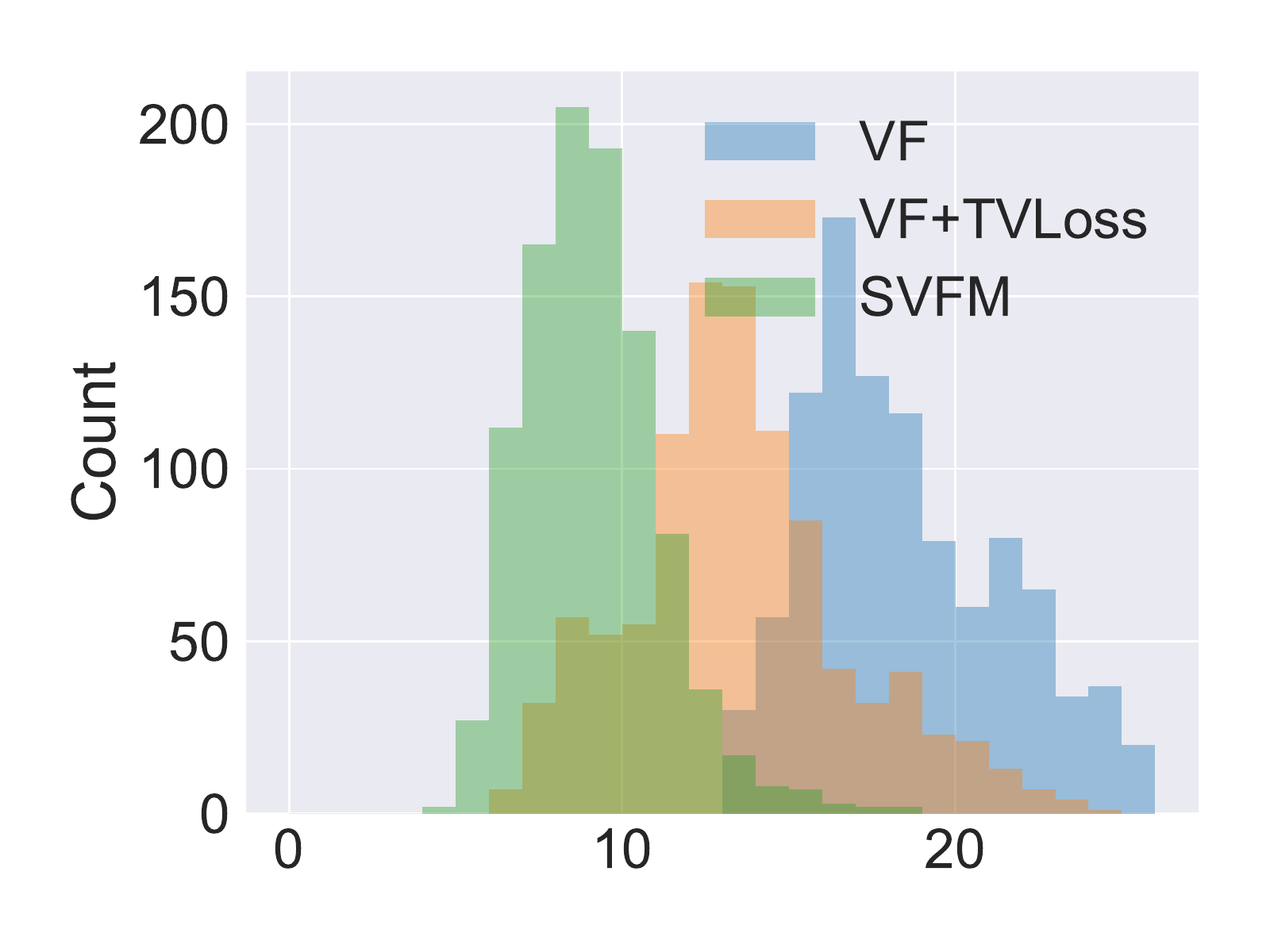}} 
    \subfigure[XOR]{\label{fig:nfe:xor}  \includegraphics[width=0.3\linewidth]{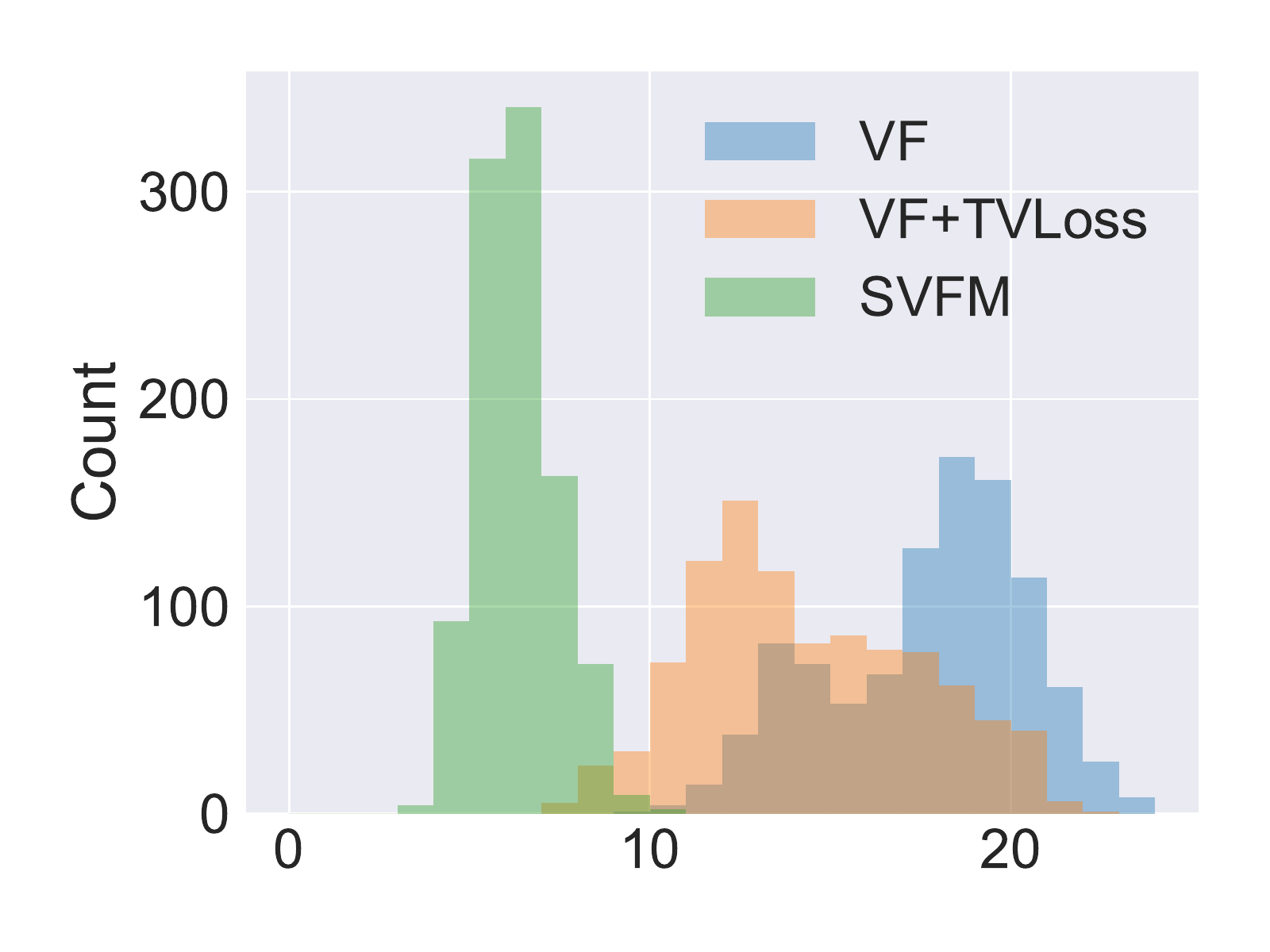}}\\
    \subfigure[VF \vs VF+TVLoss]{\label{fig:nfe:xor:vf:vfl}  \includegraphics[width=0.3\linewidth]{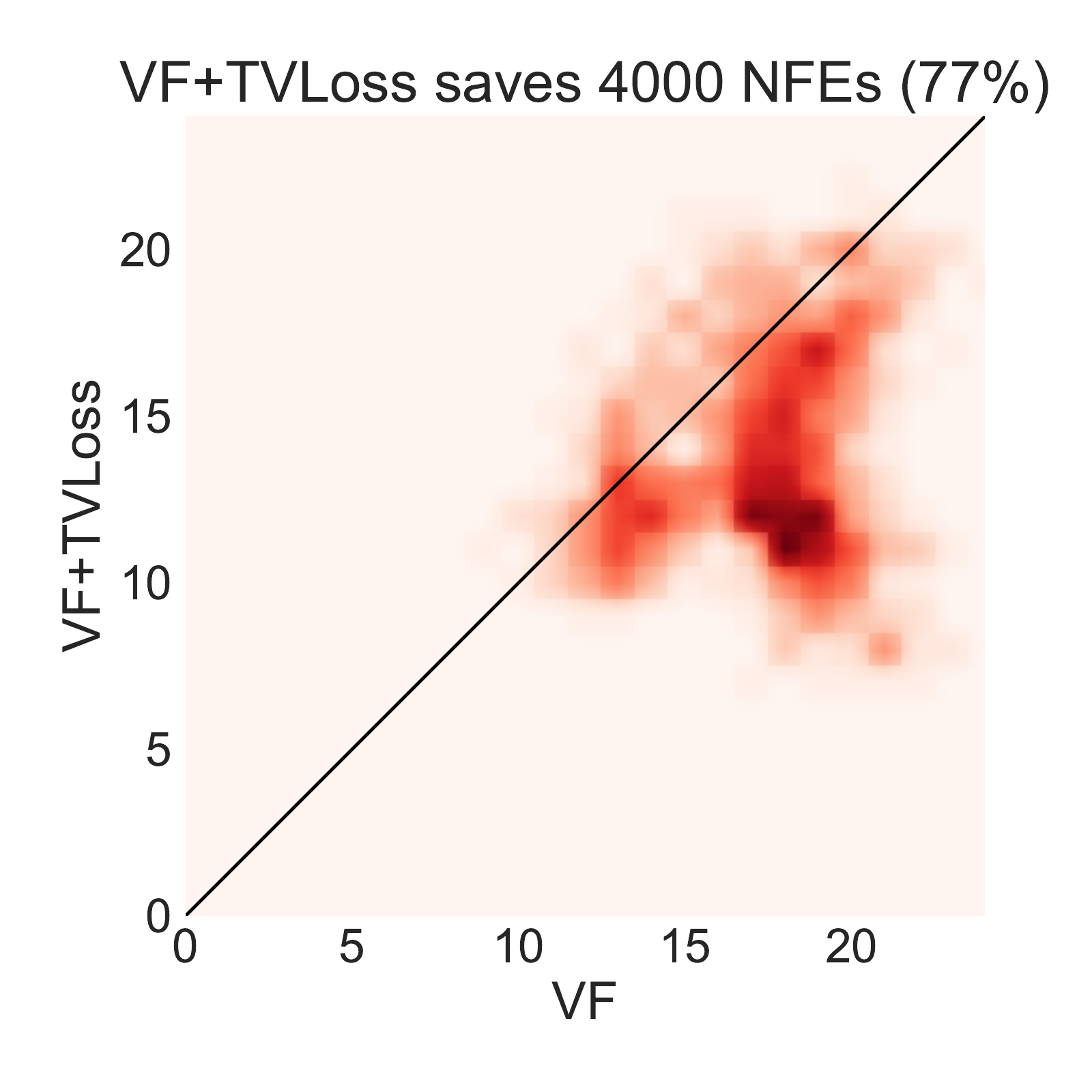}}
    \subfigure[VF \vs SVFM]{\label{fig:nfe:xor:vf:svfm} \includegraphics[width=0.3\linewidth]{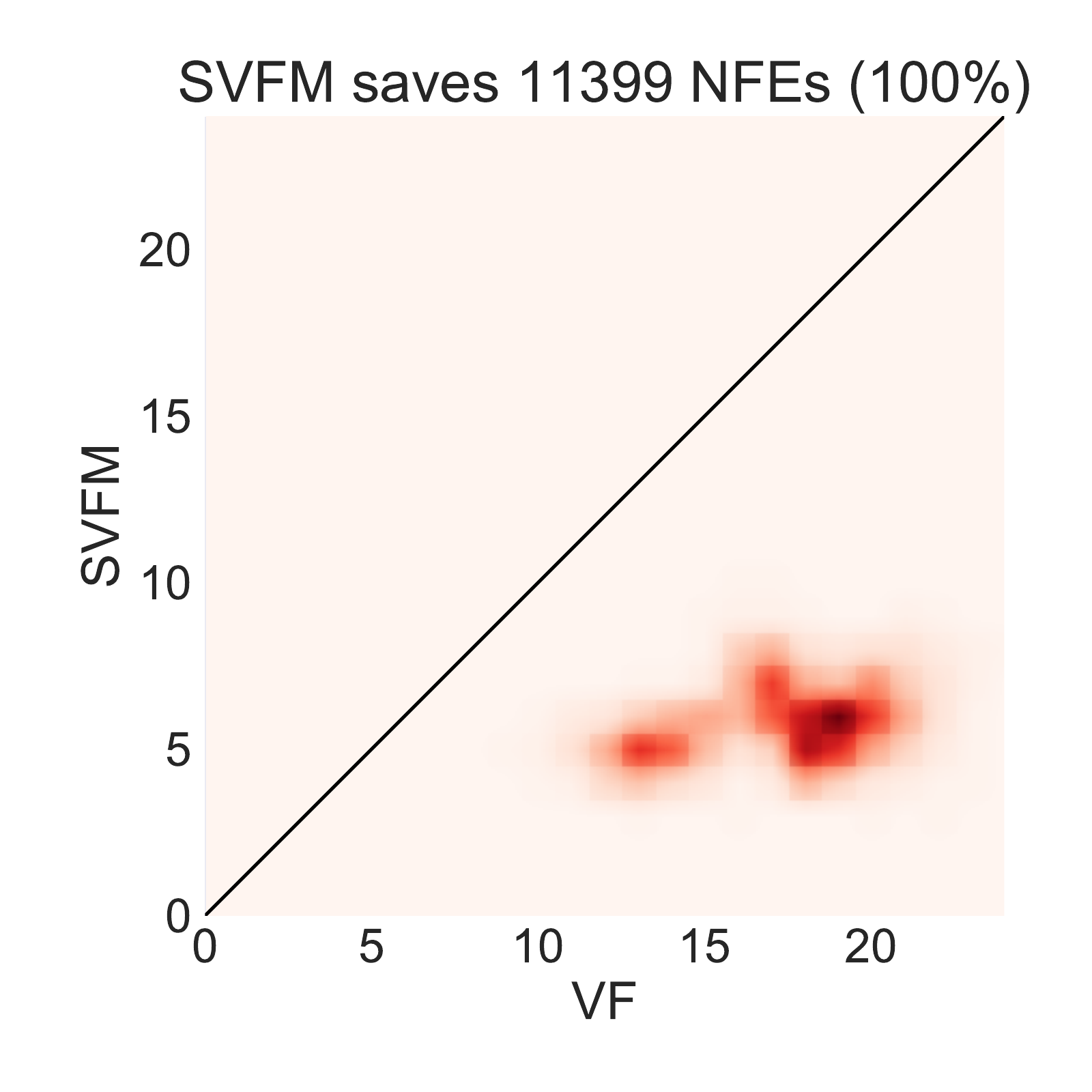}}
    \subfigure[VF+TVLoss \vs SVFM]{\label{fig:nfe:xor:vfl:svfm} \includegraphics[width=0.3\linewidth]{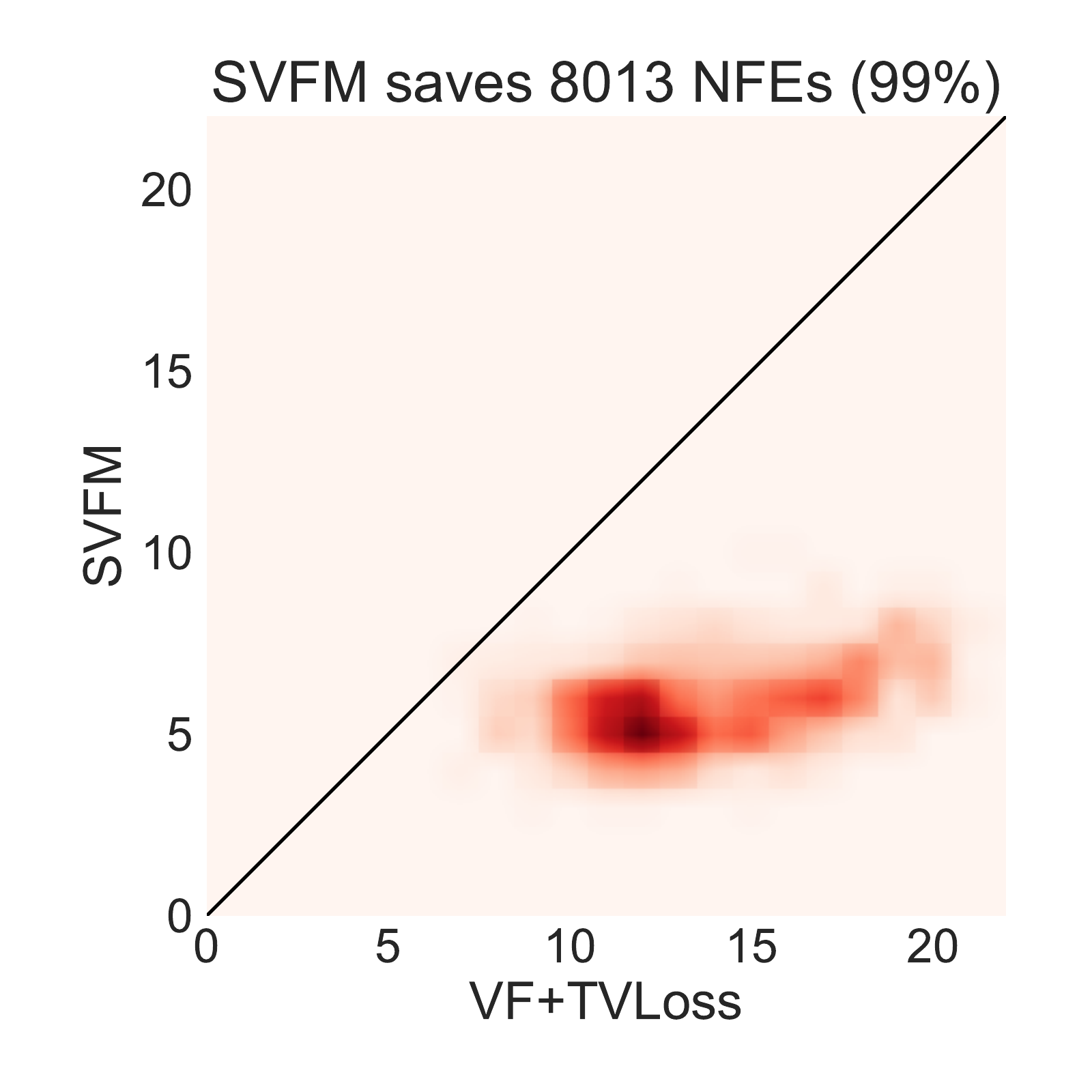}}
    \caption{NFE histograms for baseline and proposed models on moons (\cref{fig:nfe:moons}), circles (\cref{fig:nfe:nest}) and XOR (\cref{fig:nfe:xor}) dataset. The per-instance NFE heatmap is shown for the XOR dataset in \cref{fig:nfe:xor:vf:vfl,fig:nfe:xor:vf:svfm,fig:nfe:xor:vfl:svfm}. Here, red indicates regions high density and the leading diagonal is the line of equal NFEs. SVFM models consistently require fewer NFEs. The title of these figures tallies the total NFE savings and proportions of instances in receipt of these savings.}
    \label{fig:nfe_full}
\end{figure}

\textbf{Forward evaluation analysis.}
%
%
%
The effective complexity of \ac{NODE} models is widely reported with the \ac{NFE} metric, but \ac{NFE} is a somewhat nebulous term since it more accurately captures the maximal \ac{NFE} of a batch. We illustrate this by solving the \ac{IVP} for every instance of a dataset separately and producing a histogram on the set of \acp{NFE}. 
Taking the particular example of the \ac{VF} model on the moons dataset (\ie blue histogram in \cref{fig:nfe:moons}) the \ac{NFE} metric (in the traditional sense) is 26, but it is clear that most datapoints require much fewer function evaluations, with the mean and median $\approx 17$. Computational savings can be achieved by separating the `hard' and `easy' instances from one another since it is the hardest instance of a batch that subjects the rest to unnecessary evaluations. Estimation error given by embedded solvers allow for dynamic evaluation. 

The remining figures in the top row of \cref{fig:nfe_full} present the \ac{NFE} histograms on three datasets, and baseline, \ac{VF}+\ac{PLL} and \ac{SVFM} results are shown in blue, orange and green respectively. The basic \ac{NODE} models are shown here to require the most function evaluations. Although incorporating \ac{PLL} reduces the number of samples required, the greatest savings in all cases are obtained with the \ac{SVFM} model since it directly facilitates linear solutions. 

The \ac{NFE} distributions from pairs of models are compared in the lower row of \cref{fig:nfe_full} on a per-instance basis. Here, the $(i, j)$-th component of the matrix counts the number of instances for which the first model made $i$ evaluations and the second made $j$. The \acp{NFE} for \ac{VF}, \ac{VF}+\ac{PLL} and \ac{SVFM} models are compared on the \ac{XOR} dataset. This figure shows that \ac{NFE} savings are gained by using the proposed method and \ac{PLL} losses over baselines, with the biggest savings delivered by \ac{SVFM}. In particular with \cref{fig:nfe:xor:vf:svfm,fig:nfe:xor:vfl:svfm} we see that the proposed method dominates \acp{VF} in all cases. The supplement contain per-instance \ac{NFE} distributions on the other datasets investigated and, in general, show more significant savings with the \ac{SVFM} approach. 

\textbf{Behavioural forecasting.} In behavioural forecasting we are interested in predicting the future location and trajectory of a person given knowledge of their original starting position. \acp{SVFM} utility in delivering these forecasts is demonstrated with a bespoke dataset captured in a residential environment. A volunteer was recorded with a \ac{LIDAR} data collection unit and the resulting pointcloud was processed with \ac{SLAM} techniques to produce relative time and location data. The protocol is detailed in the supplement. For the purposes of this experiment, data was collected over four consistent paths uniformly that start in the living room and end at the front door, kitchen, landing and dining room. The behavioural data will be released on publication in raw and processed forms. 

\begin{figure}[t]
    \centering
    \subfigure[Forecasts learnt from uniform endpoints.]{\label{fig:behav:uniform}\includegraphics[width=0.32\linewidth,angle=-90]{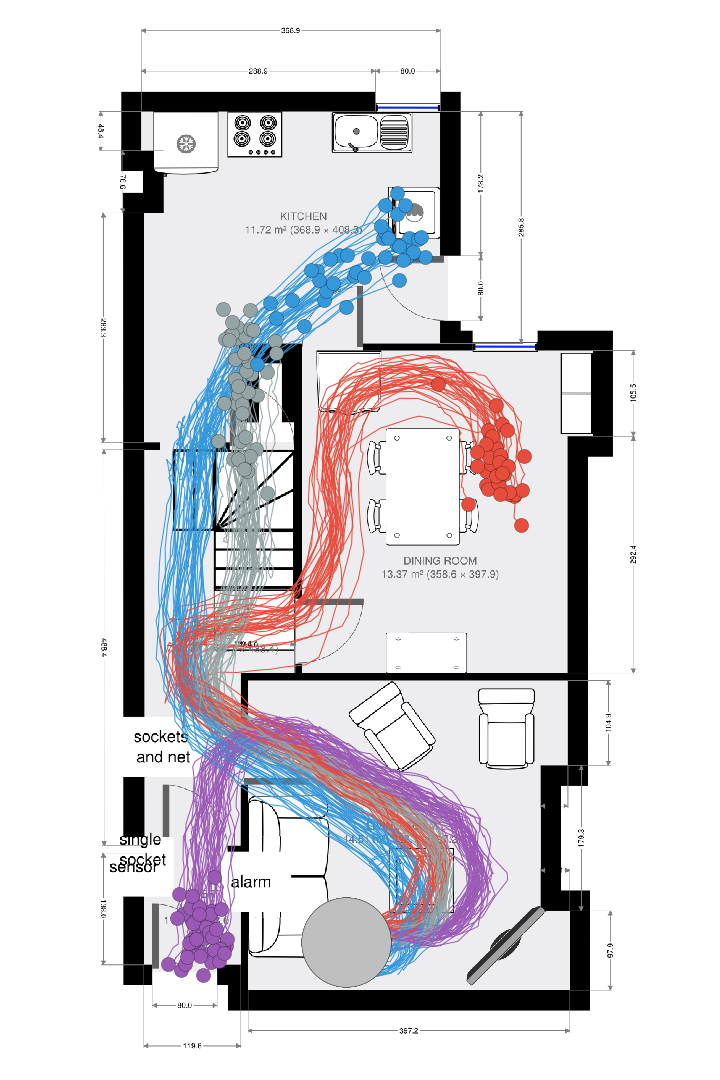}} 
    \subfigure[Conterfactual forecast based on time of day.]{\label{fig:behav:counterfact}\includegraphics[width=0.32\linewidth,angle=-90]{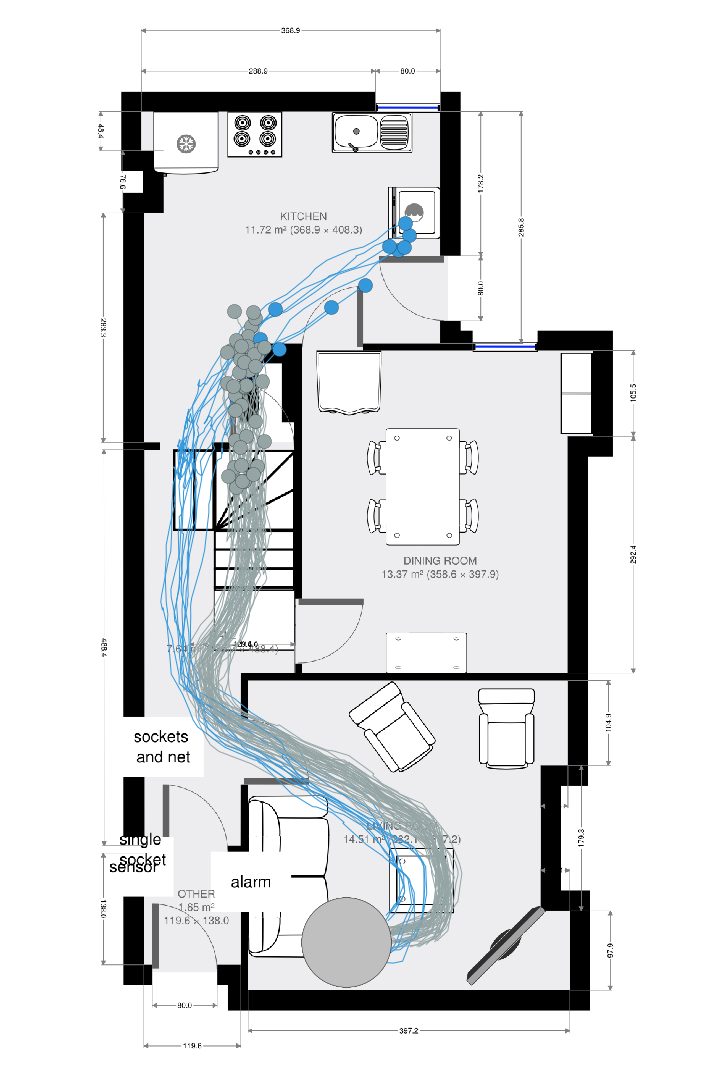}} 
    \caption{Visualisation of behavioural forecasting starting from the couch in the living room (grey circle). Purple, blue, grey and red represent front door, kitchen, landing (up stairs) and dining room trajectories respectively. }
    \vspace{-1em}
    \label{fig:behaviour}
\end{figure}

In \cref{fig:behav:uniform} the floor plan of the residential environment and the approximate location of furniture are shown. An \ac{SVFM} model is learnt on the data above with the forecasting loss (\cref{eq:apl}), and 200 paths sampled from the learnt model are overlaid on this figure. Four clear trajectories corresponding to the four endpoints mentioned above can be seen in the figure and are colour coded according to endpoint. Although the essential trajectory of the paths is the same, each particular path is unique. This is due to directional and length uncertainty. Length uncertainty is particularly notable with the kitchen trajectories (blue) in which endpoints span a region of approximately two metres. 

The distribution over endpoints was balanced in our data capture, and hence in \cref{fig:behav:uniform}. Although this clearly illustrates the model's capability in forecasting behaviour, it does not represent the dynamics of naturalistic behaviour since people will not enter rooms uniformly at random in general. In order to demonstrate the model's ability to capture naturalistic behaviour, we synthesise a dataset where the encountered paths are conditioned on the \ac{TOD}. `Day' and `night' periods are defined, and during the daytime all paths are walked with uniform probability. During `night,' however, the landing and kitchen are selected with probability 0.9 and 0.1 respectively. An \ac{SVFM} model influenced by the time tuple is learnt on these paths. We test with the same conditions of \cref{fig:behav:uniform} during the `day,' and endpoints mimic those in \cref{fig:behav:uniform}. However, a counterfactual query of the following form was made: `what would the endpoints be had the \ac{TOD} been night instead of day?' The resulting paths are shown in \cref{fig:behav:counterfact}, and the model has delivered on the expectation that `landing' endpoints should be favoured given the temporal context of the query. This is a more complete demonstration of the cyclic continuation demonstrated earlier in \cref{fig:periodic}. 

Assessment of forecasting models like this are of particular interest in healthcare domains. A growing pressure for passive assessment disease state and prognosis is leading to better characterisation of symptom such as `wandering behaviour' for patients with Alzheimer's disease. This symptom is characterised by high-entropy transit within the home, and this is exactly what is shown in \cref{fig:behav:uniform}. These are easily distinguished from the low-entropy transitions such as those shown in \cref{fig:behav:counterfact}. Baseline \ac{NODE} and \ac{ANODE} models are unable to adequately characterise these behavioural problems due to the natural variation of walked paths in these domains. 

\section{Conclusion}
\label{section:conclusion}

This paper is concerned with encouraging parsimonious advections from neural ordinary differential equation models, and we draw two main conclusions in this work that are fortified by our experiments: 1. loss functions acting on particle trajectories lead to models with lower effective complexity; and 2. introducing uncertainty to vector fields directly delivers simpler solutions. To the best of our knowledge, trajectory-based losses have not previously been explored within the \ac{NODE} paradigm, and the three that we introduce are shown here to be a straightforward and effective means of controlling model complexity. The most successful model in our experiments is the stochastic vector field mixture that we propose, and it is the vector field mixtures in particular that facilitate improved performance and linearly transported solutions. The utility of these models is demonstrated on illustrative tasks and in human behavioural modelling where we demonstrate its ability to capture the rich variation that is characteristic and forecast human movement. These results in particular suggest that these models may be applied to open behavioural quantification questions in healthcare settings, particularly with Alzheimer's disease. 
Future work will build on what is presented here, and in particular will characterise between-resident effects in forecasting and explore behavioural modelling when location is uncertain or latent. 

%
%
%

\subsubsection*{Acknowledgments}

This research was conducted under the `Continuous Behavioural Biomarkers of Cognitive Impairment' project funded by the UK Medical Research Council Momentum Awards under Grant MC/PC/16029.

{
    \small
    \bibliographystyle{unsrtnat}
    \bibliography{main}

\begin{thebibliography}{22}
\providecommand{\natexlab}[1]{#1}
\providecommand{\url}[1]{\texttt{#1}}
\expandafter\ifx\csname urlstyle\endcsname\relax
  \providecommand{\doi}[1]{doi: #1}\else
  \providecommand{\doi}{doi: \begingroup \urlstyle{rm}\Url}\fi

\bibitem[He et~al.(2016)He, Zhang, Ren, and Sun]{he2016deep}
Kaiming He, Xiangyu Zhang, Shaoqing Ren, and Jian Sun.
\newblock Deep residual learning for image recognition.
\newblock In \emph{Proceedings of the IEEE conference on computer vision and
  pattern recognition}, pages 770--778, 2016.

\bibitem[Lu et~al.(2017)Lu, Zhong, Li, and Dong]{lu2017beyond}
Yiping Lu, Aoxiao Zhong, Quanzheng Li, and Bin Dong.
\newblock Beyond finite layer neural networks: Bridging deep architectures and
  numerical differential equations.
\newblock \emph{arXiv preprint arXiv:1710.10121}, 2017.

\bibitem[Weinan(2017)]{weinan2017proposal}
E~Weinan.
\newblock A proposal on machine learning via dynamical systems.
\newblock \emph{Communications in Mathematics and Statistics}, 5\penalty0
  (1):\penalty0 1--11, 2017.

\bibitem[Ruthotto and Haber(2018)]{ruthotto2018deep}
Lars Ruthotto and Eldad Haber.
\newblock Deep neural networks motivated by partial differential equations.
\newblock \emph{arXiv preprint arXiv:1804.04272}, 2018.

\bibitem[Chen et~al.(2018)Chen, Rubanova, Bettencourt, and
  Duvenaud]{chen2018neural}
Tian~Qi Chen, Yulia Rubanova, Jesse Bettencourt, and David~K Duvenaud.
\newblock {Neural ordinary differential equations}.
\newblock In \emph{Advances in Neural Information Processing Systems}, pages
  6572--6583, 2018.

\bibitem[Dupont et~al.(2019)Dupont, Doucet, and Teh]{dupont2019augmented}
Emilien Dupont, Arnaud Doucet, and Yee~Whye Teh.
\newblock {Augmented Neural ODEs}.
\newblock \emph{arXiv preprint arXiv:1904.01681}, 2019.

\bibitem[Minka and Winn(2009)]{minka2009gates}
Tom Minka and John Winn.
\newblock Gates.
\newblock In \emph{Advances in Neural Information Processing Systems}, pages
  1073--1080, 2009.

\bibitem[Rabiner and Juang(1986)]{rabiner1986introduction}
Lawrence~R Rabiner and Biing-Hwang Juang.
\newblock An introduction to hidden markov models.
\newblock \emph{ieee assp magazine}, 3\penalty0 (1):\penalty0 4--16, 1986.

\bibitem[Sutton et~al.(2012)Sutton, McCallum, et~al.]{sutton2012introduction}
Charles Sutton, Andrew McCallum, et~al.
\newblock An introduction to conditional random fields.
\newblock \emph{Foundations and Trends{\textregistered} in Machine Learning},
  4\penalty0 (4):\penalty0 267--373, 2012.

\bibitem[Zheng et~al.(2015)Zheng, Jayasumana, Romera-Paredes, Vineet, Su, Du,
  Huang, and Torr]{zheng2015conditional}
Shuai Zheng, Sadeep Jayasumana, Bernardino Romera-Paredes, Vibhav Vineet,
  Zhizhong Su, Dalong Du, Chang Huang, and Philip~HS Torr.
\newblock Conditional random fields as recurrent neural networks.
\newblock In \emph{Proceedings of the IEEE international conference on computer
  vision}, pages 1529--1537, 2015.

\bibitem[Otto et~al.(2010)Otto, Germer, Hege, and Theisel]{otto2010uncertain}
Mathias Otto, Tobias Germer, Hans-Christian Hege, and Holger Theisel.
\newblock Uncertain 2d vector field topology.
\newblock In \emph{Computer Graphics Forum}, volume~29, pages 347--356. Wiley
  Online Library, 2010.

\bibitem[Bishop(1994)]{bishop1994mixture}
Christopher~M Bishop.
\newblock Mixture density networks.
\newblock Technical report, Citeseer, 1994.

\bibitem[Kutta(1901)]{kutta1901beitrag}
Wilhelm Kutta.
\newblock Beitrag zur n{\"a}herungweisen integration totaler
  differentialgleichungen.
\newblock 1901.

\bibitem[Dormand and Prince(1980)]{dormand1980family}
John~R Dormand and Peter~J Prince.
\newblock A family of embedded runge-kutta formulae.
\newblock \emph{Journal of computational and applied mathematics}, 6\penalty0
  (1):\penalty0 19--26, 1980.

\bibitem[S{\"a}rkk{\"a} and Solin(2019)]{sarkka2019applied}
Simo S{\"a}rkk{\"a} and Arno Solin.
\newblock \emph{Applied Stochastic Differential Equations}, volume~10.
\newblock Cambridge University Press, 2019.

\bibitem[Ruiz et~al.(2016)Ruiz, AUEB, and Blei]{ruiz2016generalized}
Francisco~R Ruiz, Michalis Titsias~RC AUEB, and David Blei.
\newblock The generalized reparameterization gradient.
\newblock In \emph{Advances in neural information processing systems}, pages
  460--468, 2016.

\bibitem[Jang et~al.(2016)Jang, Gu, and Poole]{jang2016categorical}
Eric Jang, Shixiang Gu, and Ben Poole.
\newblock Categorical reparameterization with gumbel-softmax.
\newblock \emph{arXiv preprint arXiv:1611.01144}, 2016.

\bibitem[Howard(1998)]{howard1998gronwall}
Ralph Howard.
\newblock The gronwall inequality.
\newblock \emph{lecture notes}, 1998.

\bibitem[Butcher and Goodwin(2008)]{butcher2008numerical}
John~Charles Butcher and Nicolette Goodwin.
\newblock \emph{Numerical methods for ordinary differential equations},
  volume~2.
\newblock Wiley Online Library, 2008.

\bibitem[Kingma and Ba(2014)]{kingma2014adam}
Diederik~P Kingma and Jimmy Ba.
\newblock Adam: A method for stochastic optimization.
\newblock \emph{arXiv preprint arXiv:1412.6980}, 2014.

\bibitem[Hess et~al.(2016)Hess, Kohler, Rapp, and Andor]{hess2016real}
Wolfgang Hess, Damon Kohler, Holger Rapp, and Daniel Andor.
\newblock Real-time loop closure in 2d lidar slam.
\newblock In \emph{Robotics and Automation (ICRA), 2016 IEEE International
  Conference on}, pages 1271--1278. IEEE, 2016.

\bibitem[Thrun et~al.(2005)Thrun, Burgard, and Fox]{thrun_probabilistic_2005}
S.~Thrun, W.~Burgard, and D.~Fox.
\newblock \emph{Probabilistic {Robotics} ({Intelligent} {Robotics} and
  {Autonomous} {Agents})}.
\newblock The MIT Press, 2005.
\newblock ISBN 978-0-262-20162-9.

\end{thebibliography}
}

\begin{appendices}

\section{Theoretical analysis}


\subsection{NODE and ANODE fail cases} 

In this section we introduce some theoretical discussions on \acp{ODE} and \acp{NODE} in solving crossing, splitting and scaling problems. We draw from existing work that shows \acp{NODE} cannot solve crossing and also demonstrate that neither \acp{NODE} nor \acp{ANODE} can solve splitting and scaling. 

The following notation is used: $\h(t_T) \in \mathbb{R}^D$ represents the solution of the \ac{IVP} problem with vector fields defined as $\nabla \h(t) = f(\h(t), t; \thetab) \in \mathbb{R}^D$ over the integration interval $[0, T]$. $\h_i(t)$ indicates the transformation at $t$ of a particular instance, $i$. We set up by established standard conventions by omitting the parameterisation of $\thetab$ on $f$ for notational convenience and $f$ is assumed to be Lipschitz continuous, \ie $\forall t$ there exists a positive $C$ such that

\begin{align}
    \| f(\h_1(t), t) - f(\h_2(t), t) \| \leq C \| \h_1(t) - \h_2(t)  \|.
\end{align}




\paragraph{Proposition.} \textit{Let $\h_1(T)$ and $\h_2(T)$ be two solutions of an \ac{ODE} with different initial conditions, \ie $\h_1(0) \neq \h_2(0)$. Then, for all $t \in (0, T]$, $\h_1(t) \neq \h_2(t)$. }

\textit{Proof.} Proof follows naturally by contradiction based on the existence and uniqueness of \ac{ODE} solutions. If there exists some $t^* \in (0, T]$ for which $\h_1(t^*) = \h_2(t^*)$, then let $\h_3(t^*)$ be a new \ac{IVP} defined at $t^*$. Since existence and uniqueness of \acp{ODE} hold when solving \acp{IVP} forward and backward, the backward solution $\h_3(0)$ is unique. This contradicts the setup where $\h_1(0)$ and $\h_2(0)$ are distinct, and thus the proposition is true. 

This proposition is useful in this setting since it demonstrates that \ac{ODE} solutions can never intersect because otherwise the fundamental existence and uniqueness properties of \acp{ODE} are violated. \acp{ODE}, therefore, have no solutions for crossing, splitting and scaling problems. 

This theory is generalised to \acp{NODE} by \citet{dupont2019augmented}. We do not duplicate the proof here, but assuming that $f$ is Lipschitz continuous they show that the feature mappings are homeomorphisms and makes use of Gronwall's lemma \cite{howard1998gronwall} in the proof. With this established, it is trivial to generalise corollaries that demonstrate that crossing, splitting and scaling problems cannot be solved by \acp{NODE}. 

Augmented \acp{NODE} were introduced to solve the crossing problem by concatenating the data with zeros and solving the \ac{IVP} with this new augmented representation. \acp{ANODE} flows may use the additional dimension(s) in finding solution paths which provide a solution for the crossing problem. Let us formally analyse \ac{NODE}'s ability to tackle splitting and scaling. Let $g_s: \mathbb{R} \rightarrow \mathbb{R}$ be a function such that 

\begin{align}
    \label{eq:splitfunc}
    g_s = \begin{cases}
         \text{-1 with probability 0.5} \\
         \text{\phantom{-}1 with probability 0.5}
    \end{cases}
\end{align}

\paragraph{Proposition.} \textit{\ac{ANODE} flows cannot model $g_s$.}

\textit{Proof.} The proof of this follows directly from the fact that \acp{ANODE} retain the theoretical properties of \acp{NODE} and thus their feature mappings are homeomorphisms \cite{dupont2019augmented}. Therefore, they do not solve $g_s$ since it is a one-to-many mapping.

A proof for the scaling function follows naturally with the same rationale. The key problems is that since the initial conditions of the data are identical, their paths will remain the same (even with augmentation) and will never become untangled. Our proposed model overcomes this limitation by assigning data to a vector field probabilistically which allows \cref{eq:splitfunc} to be modelled.

\subsection{VF variance and alignment}

Aligned \acp{VF} are encouraged by the variance reduction loss discussed in the main text, and in this section we expand on the computational gains achieved when solving \acp{IVP} with aligned \acp{VF}.  

We formalise this analysis with Runge-Kutta-based iterative embedded \ac{ODE} solvers. These can be characterised by their Butcher tableau \cite{butcher2008numerical} that defines the coefficients, temporal offsets and evaluation weighting of \acp{VF}. The Butcher tableau of an order $s$ embedded solver is often seen visually as

$\hspace{2em}\begin{array}{c|cccc}
c_1    & a_{11} & a_{12}& \dots & a_{1s}\\
c_2    & a_{21} & a_{22}& \dots & a_{2s}\\
\vdots & \vdots & \vdots& \ddots& \vdots\\
c_s    & a_{s1} & a_{s2}& \dots & a_{ss} \\
\hline
       & b_1    & b_2   & \dots & b_s\\
       & b_1^*  & b_2^* & \dots & b_s^*\\
\end{array}$

and for a given state, $\h(t_{n})$, the next output is given by 

\begin{align}
    \h(t_{n + 1}) = \h(t_{n}) + h \sum _{i=1}^s b_i \bfk_i
\end{align}

where the intermediate vector field evaluations are defined as

\begin{align}
    \bfk_i = f\left(\h(t_{n}) + \sum_{j=1}^s a_{ij} \bfk_j,  t_n + c_i h \right)
\end{align}

and $a$, $b$ and $c$ are as defined in the tableau above and $h$ is the step size for that iteration. The final row of the tableau is used to estimate truncation error as follows

\begin{align}
    \h^*(t_{n+1}) &= \h(t_n) + h\sum_{i=1}^s b^*_i \bfk_i \\
    e_{t_{n+1}} &= \h(t_{n+1}) - \h^*(t_{n+1}) \nonumber \\
                &= h\sum_{i=1}^s (b_i - b^*_i) \bfk_i \label{eq:truncerr}
\end{align}

and the step size $h$ is adjusted dynamically when deriving the solution to be close to a specified tolerance value. If the error is too high, $h$ is reduced and the $(n+1)$-th iteration is repeated. If the error below the tolerance, $\h(t_{n+1})$ is accepted but the step size may be increased if the error is too small. 

If the modelling framework successfully aligns all \acp{VF} then the error term becomes

\begin{align}
    \label{eq:heaven}
    e(t_{n + 1}) &= \bfk_* \, h \,\sum_{i=1}^s (b_i - b^*_i) 
\end{align}

where $\bfk_*$ is the aligned direction and the truncation error has become independent of the intermediate evaluation points. By design $\sum_i b_i = 1$ and $\sum_i b^*_i = 1$ and so the truncation error tends to 0 in the equation above. This allows solutions to be discovered with just one step. Although the systems characterised by \cref{eq:heaven} are not particularly interesting in general, we introduce it since it is minimal in the sense of \ac{NFE} requirements. In our experimentation with variance reduction we observe significant savings using these losses in our evaluation, which can be seen to tend towards this limiting case. Interestingly, this suggests a link between transportation and variance losses since under the conditions of \cref{eq:heaven} both are minimal. 






\section{Experimental details}

\subsection{Models}
Several architectural and optimisation parameters must be specified in all experiments. These are listed here:

\begin{itemize}
    \item \ac{VF} representer
    \begin{itemize}
        \item Number of hidden layers: 1 and 2.
        \item Number of hidden units: 32 and 64 (originally 16 was considered, but it was omitted since it was never selected).
        \item Activation function: rectified units in all cases.
        \item Output activation: linear units.
    \end{itemize}
    \item Optimisation
    \begin{itemize}
        \item Optimizer: Adam \cite{kingma2014adam}.
        \item Batch size: 50 and 100.
        \item Learning rate: $10^{-2}$ and $10^{-3}$.
    \end{itemize}
    \item \ac{ODE} solver
    \begin{itemize}
        \item Tolerance: $10^{-6}$.
    \end{itemize}
    \item \ac{SVFM}
    \begin{itemize}
        \item Component selection: pick and stick, forward filtering.
    \end{itemize}
\end{itemize}

We introduced four loss functions in the main paper: \acf{MDL}, \acf{TL}, \ac{DV} and \acf{APL}. Since \ac{APL} penalises deviation from a path, it is incompatible with \ac{TL} and \ac{DV}. Classification tasks, however, may mix \ac{TL} and \ac{DV} losses with predictive losses (\eg cross-entropy). The balance between the predictive loss and the path regularisation is established with one parameter $\lambda$, and this is selected on the validation set.

\subsection{Data}

\subsubsection{Synthetic data} 

Three example datasets are used in this paper: moons, nested circles and XOR. In all cases 1,000 datapoints are sampled for training, testing and validation. The top row of \cref{fig:advection_illustration} illustrates their configuration with the positive (red) and negative (blue) classes.

\subsubsection{Behavioural data collection and processing}
The behavioural dataset was collected in an adapted residential house. The collection mechanism involved a bespoke robotic wearable, capable of gathering 2-dimensional LiDAR point clouds from a human participant. The experimental procedure asked the participant to walk naturally into 4 different sections of the house from the same origin point, located in the living room. The 4 different targets in the house included the front door, the study room, the kitchen, and the landing of the stairs. Time of each experiment was assured through NTP synchronisation between the wearable and university NTP servers.

The procedure for collection followed a protocol:

\begin{enumerate}
    \item Begin in the origin `anchor'.
    \item Choose one of the four target locations. 
    \item Proceed to the target. This should take roughly 5-10 seconds.
    \item Repeat 10 times.
\end{enumerate}

The subsequent processing of the point cloud was performed using 2-dimensional SLAM. Using MATLAB software, the data was associated using a loop closure method outlined in \cite{hess2016real}. The map and the locations were extracted using an occupancy grid map \cite{thrun_probabilistic_2005}. The loop closure in this context specifies a method of location error minimisation from aggregated sub maps \cite{hess2016real}. The parameters of the algorithm included laser reflectivity, internal loop closure threshold, resolution of the grid map and down-sampling of the data.

\begin{figure}[t]
    \centering
    \subfigure[Door]   {\label{fig:slam:d}\scalebox{-1}[1]{\includegraphics[width=0.35\textwidth,angle=-90,origin=c]{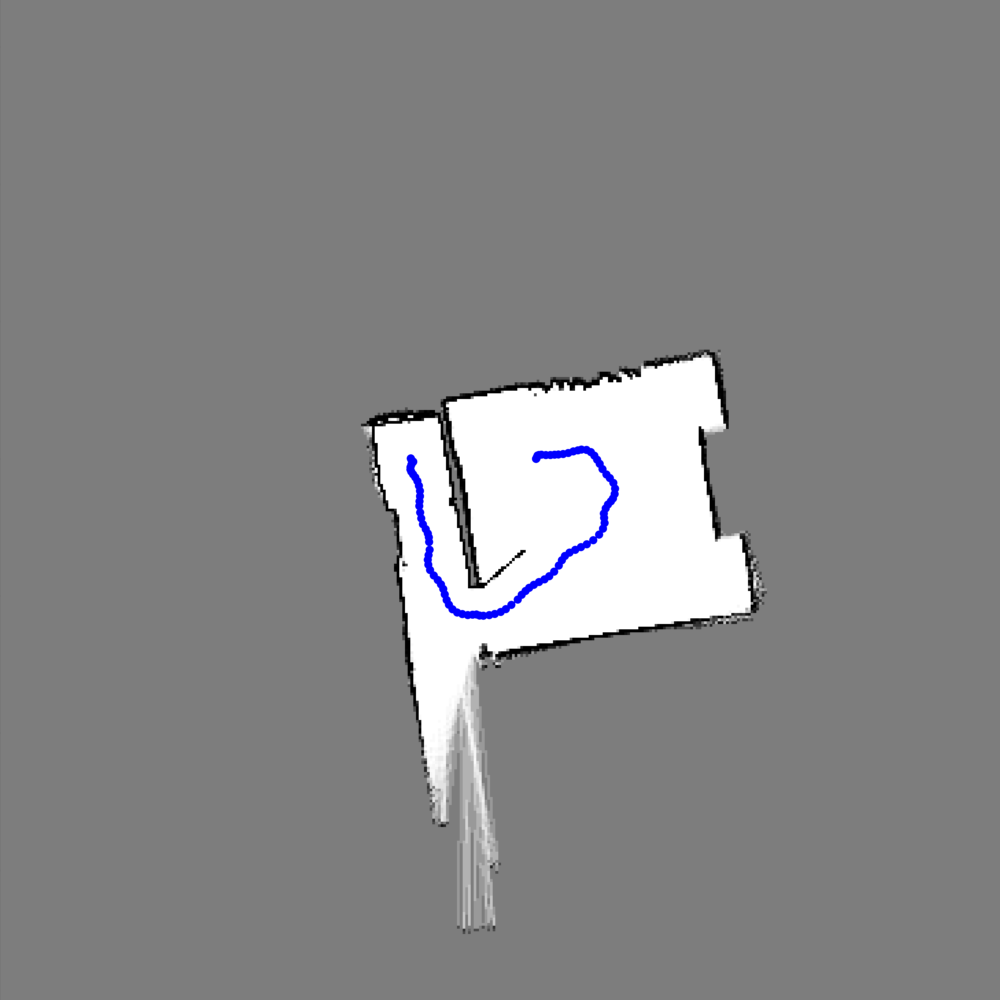}}} ~
    \subfigure[Kitchen]{\label{fig:slam:k}\scalebox{-1}[1]{\includegraphics[width=0.35\textwidth,angle=-90,origin=c]{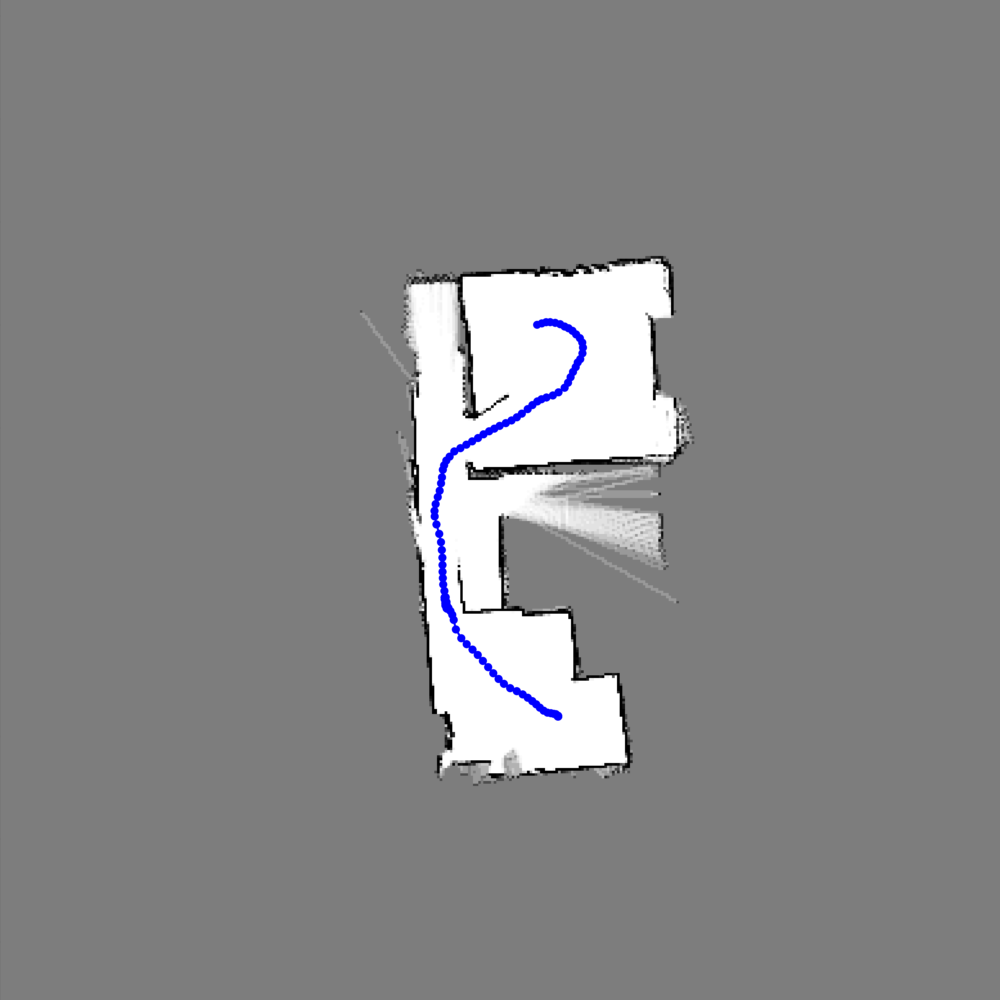}}} \\
    \subfigure[Landing]{\label{fig:slam:l}\scalebox{-1}[1]{\includegraphics[width=0.35\textwidth,angle=-90,origin=c]{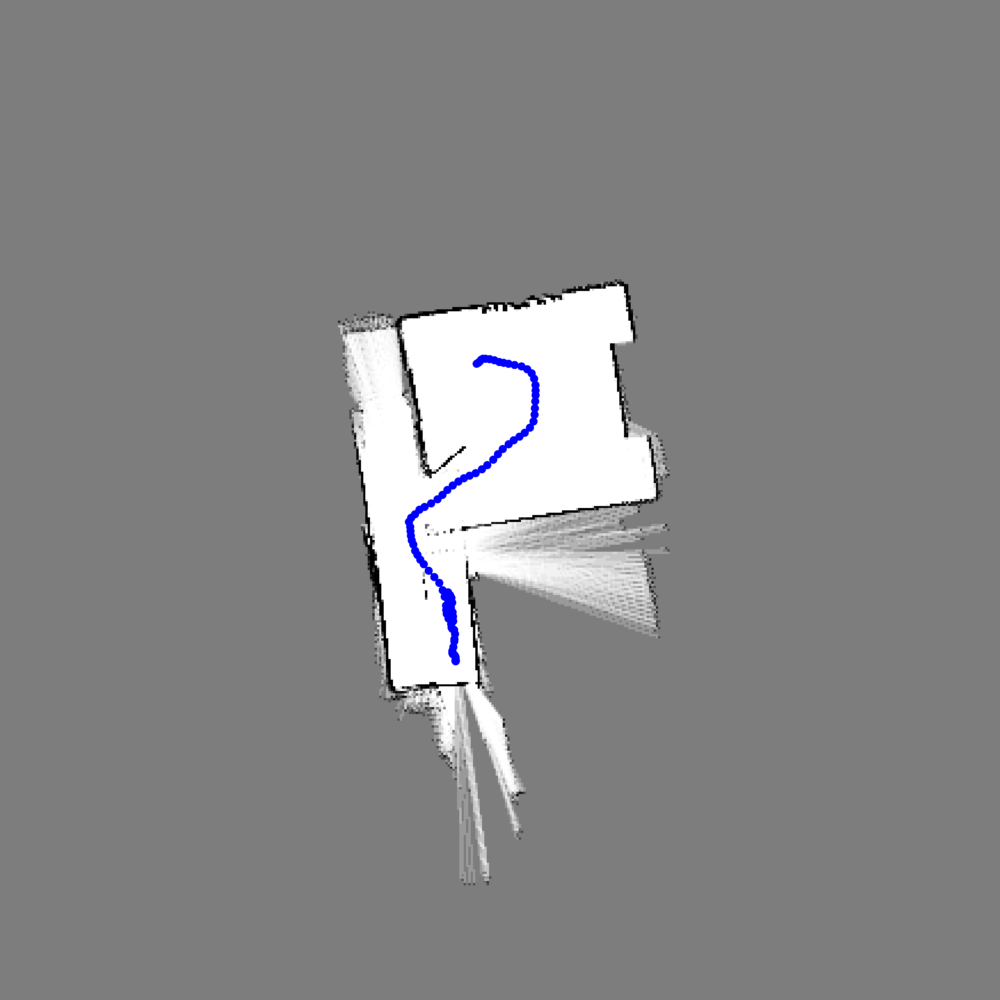}}} ~
    \subfigure[Study]  {\label{fig:slam:s}\scalebox{-1}[1]{\includegraphics[width=0.35\textwidth,angle=-90,origin=c]{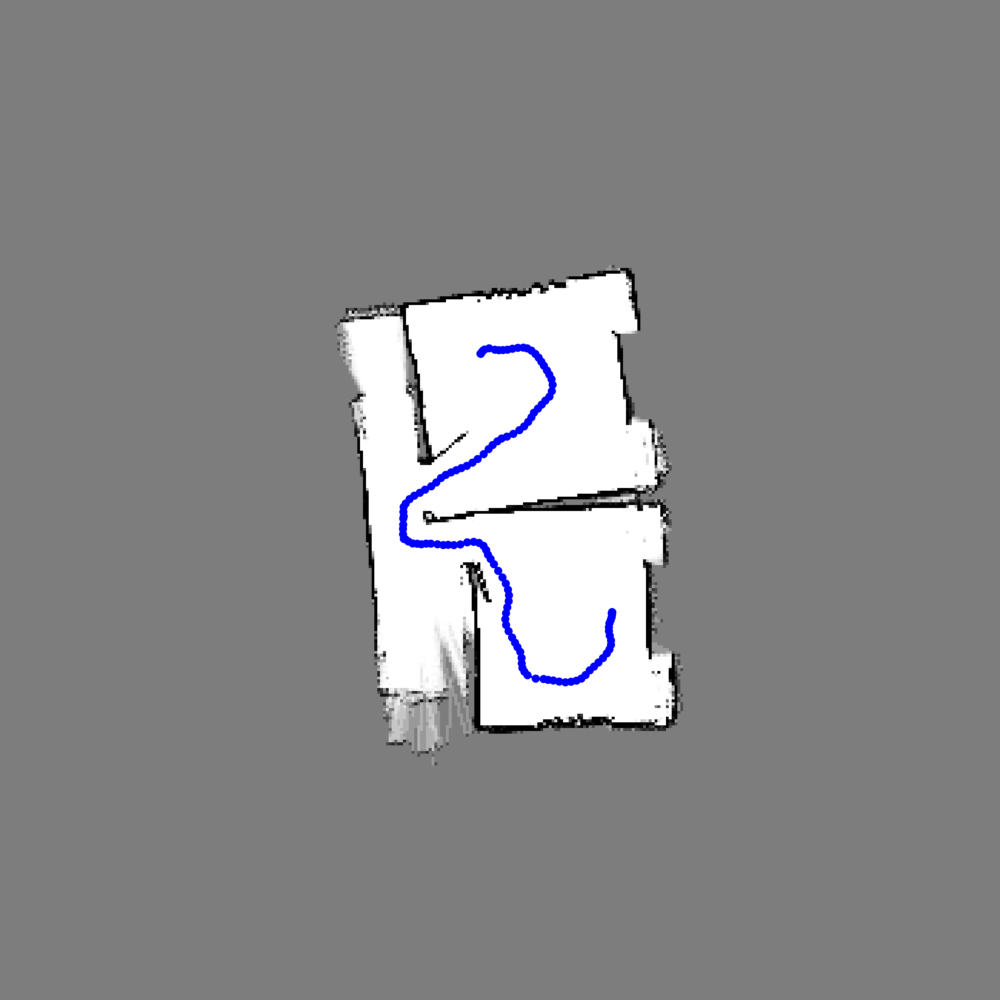}}} 
    \caption{Visualisation of SLAM maps mapping a single trajectory from the living room to the four endpoints considered in this work. We can see that the map derived from each setting is a faithful representation of the floorplan shown in the main paper.}
    \label{fig:slam_example}
\end{figure}

Examples of the produced maps, for all 4 targets can be seen in \cref{fig:slam_example}. The locations extracted using SLAM were required to be additionally associated together using Iterative Closest Point (ICP). This was done to make sure that the extracted walking tracks overlapped. The extracted location information further served as the input to the model. Some of the variation at the beginning of the paths in figs. 10 (a) and 10(b) are due to the measurement uncertainty from the low-cost LiDAR and SLAM processing. 

The raw and processed data will be made available on a public repository after publication.

\section{Supplementary results}

\begin{figure}[hbtp]
    \centering
    \subfigure{\includegraphics[width=0.3\textwidth]{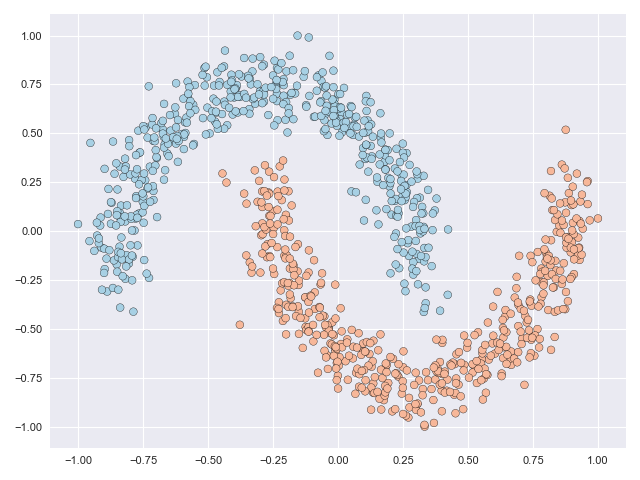}} ~
    \subfigure{\includegraphics[width=0.3\textwidth]{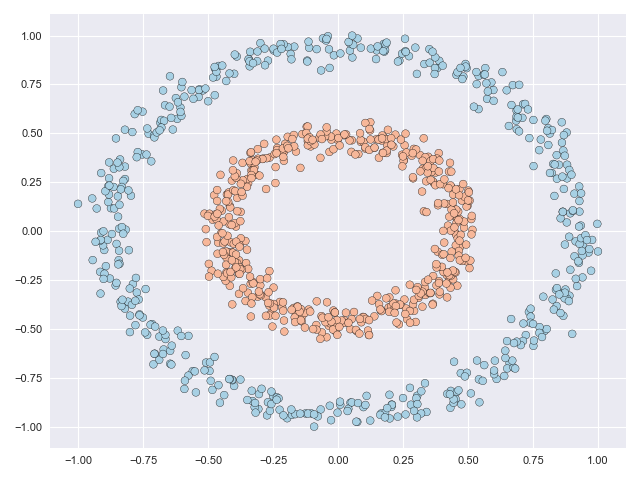}} ~
    \subfigure{\includegraphics[width=0.3\textwidth]{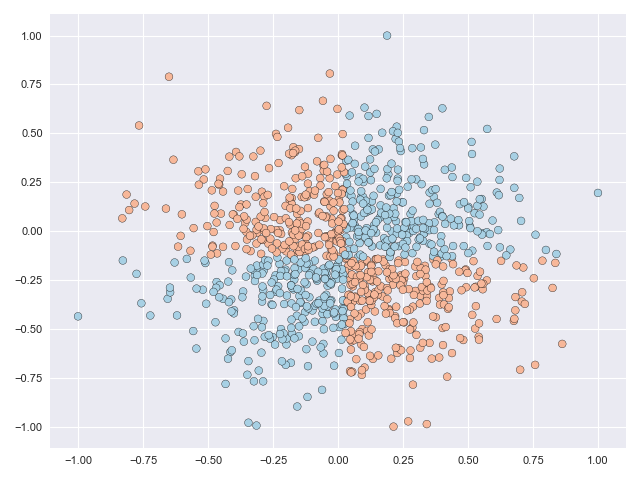}}
    \\
    \subfigure{\includegraphics[width=0.3\textwidth]{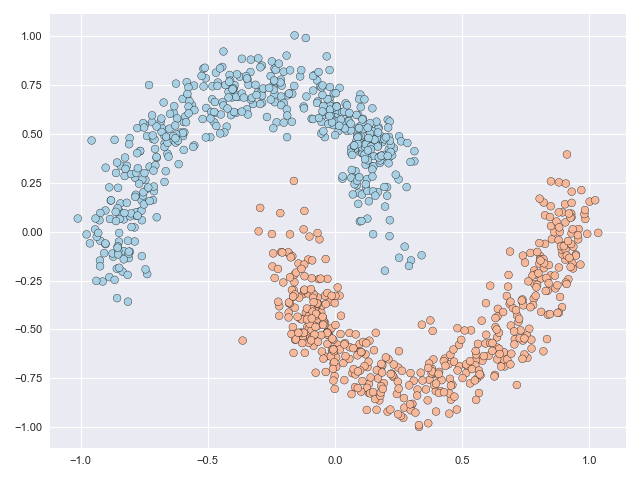}} ~
    \subfigure{\includegraphics[width=0.3\textwidth]{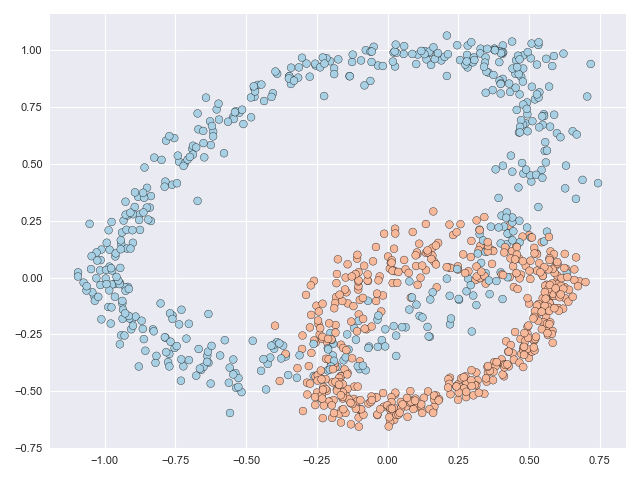}} ~
    \subfigure{\includegraphics[width=0.3\textwidth]{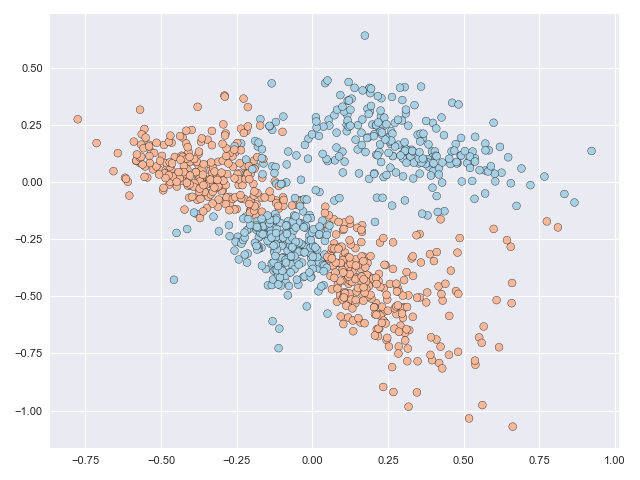}}
    \\
    \subfigure{\includegraphics[width=0.3\textwidth]{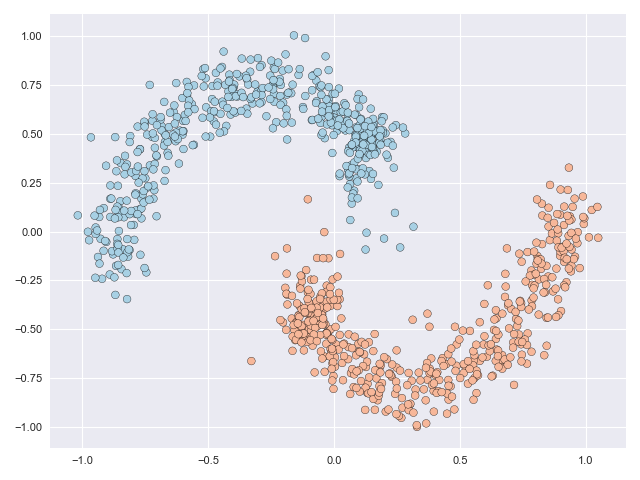}} ~
    \subfigure{\includegraphics[width=0.3\textwidth]{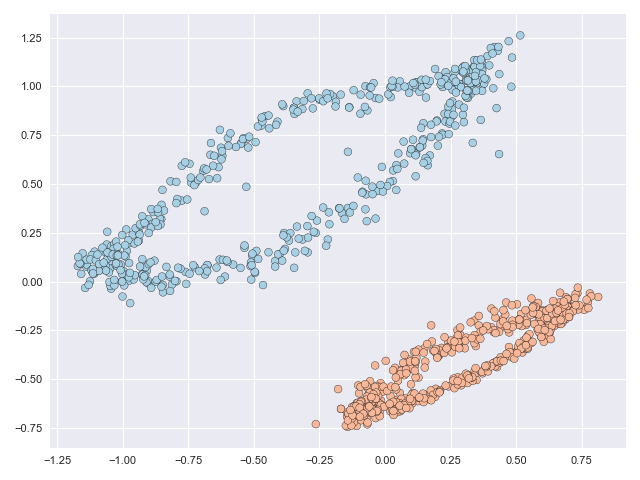}} ~
    \subfigure{\includegraphics[width=0.3\textwidth]{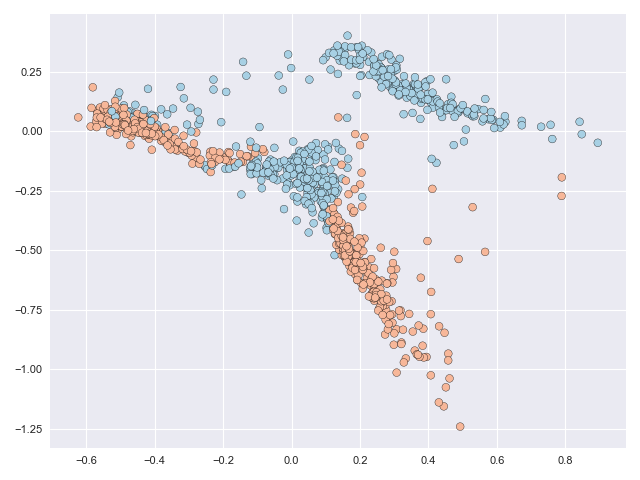}}
    \\
    \subfigure{\includegraphics[width=0.3\textwidth]{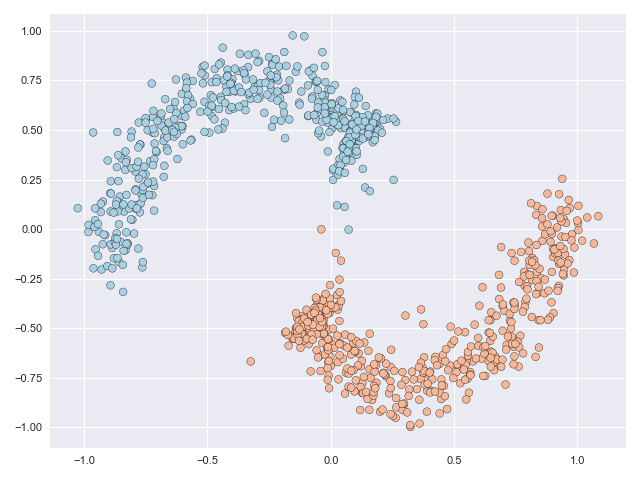}} ~
    \subfigure{\includegraphics[width=0.3\textwidth]{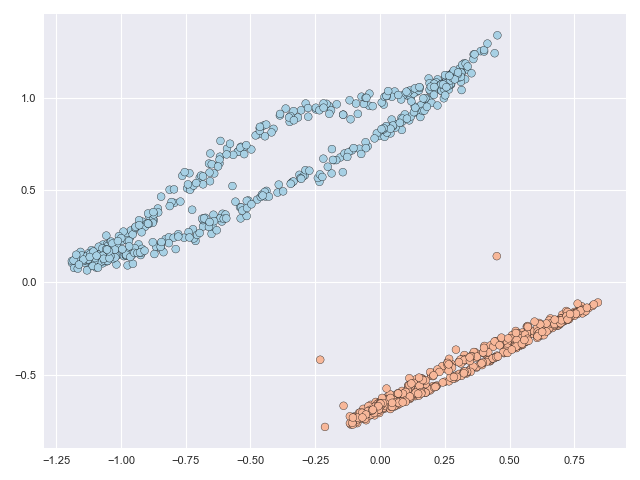}} ~
    \subfigure{\includegraphics[width=0.3\textwidth]{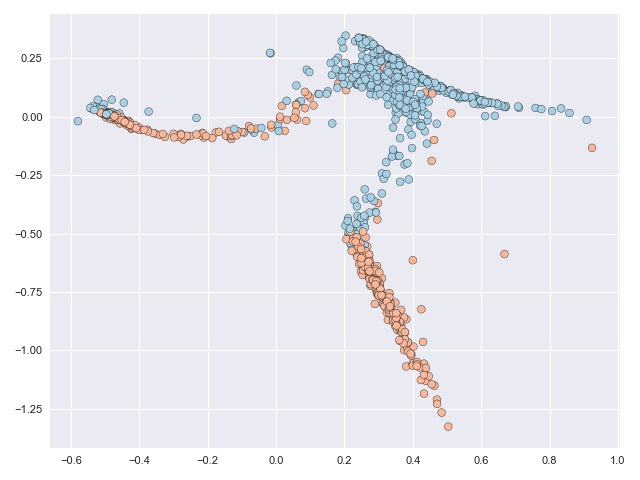}}
    \\
    \subfigure{\includegraphics[width=0.3\textwidth]{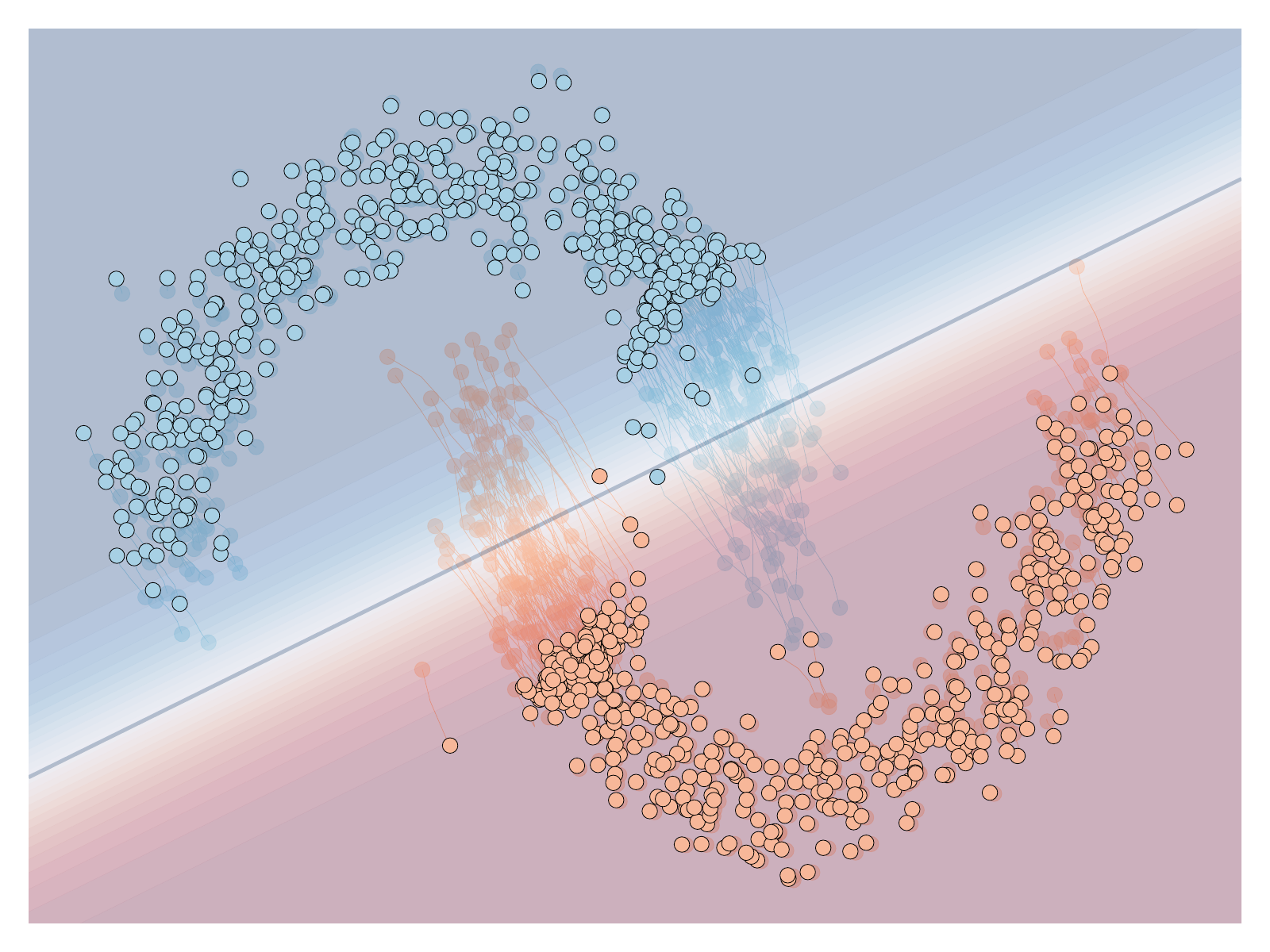}} ~
    \subfigure{\includegraphics[width=0.3\textwidth]{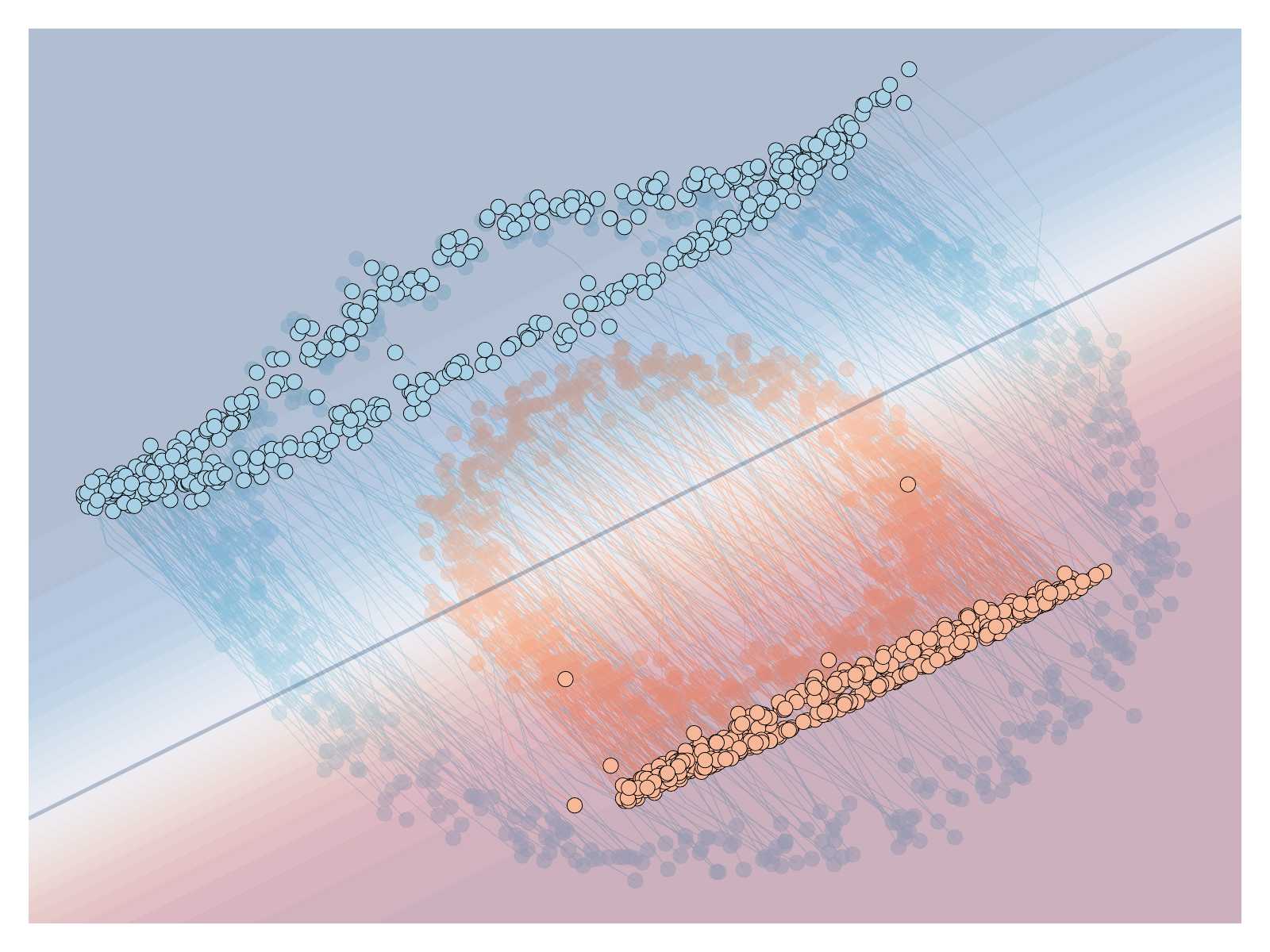}} ~
    \subfigure{\includegraphics[width=0.3\textwidth]{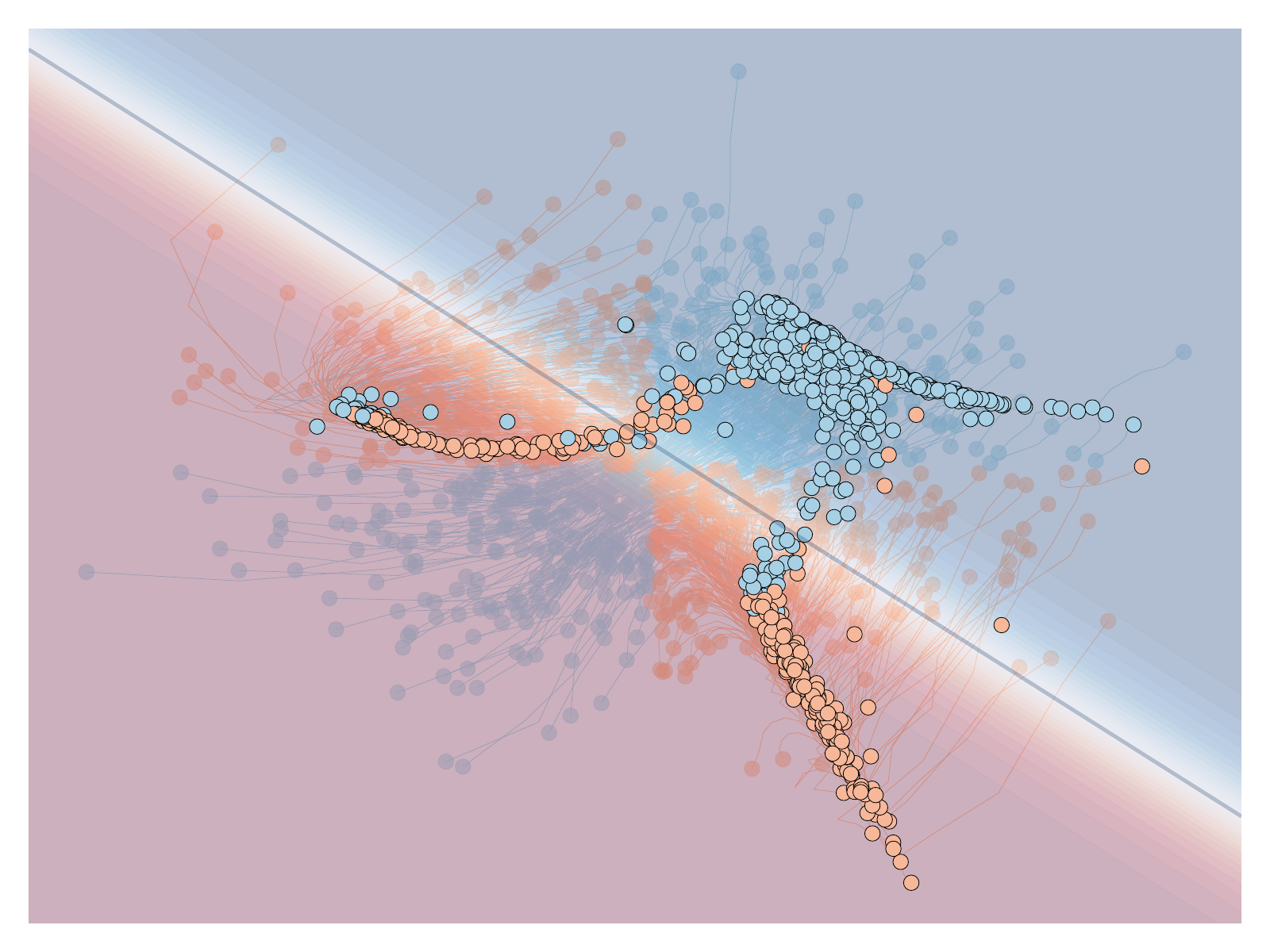}}
    \caption{Visualisation of SVFM solutions on the moons (left column), nested circle (middle column) and XOR (right column). Each row depicts a point through the solution path, and the bottom row shows the predictive distribution on the final transformation. Notice that the total particle movement of the moons solution is minimal. In the second row of the nested circles dataset the position of two classes overlap but in the next row we see that have split apart. This is because the two classes have been assigned to different vector fields. The vector field mixture helps the XOR dataset by learning to assigning blue quadrant on the top right to one component and the rest to the other. This quadrant is free to move upwards independently from the rest. The remaining clusters are assigned to the same VF and have learnt to split apart so that the lower left quadrant may join the first. Best viewed on a digital device.}
    \label{fig:advection_illustration}
\end{figure}

\begin{figure}[hbtp]
    \centering
    \subfigure[Moons VF \vs VF+TVL]  {\includegraphics[width=0.3\textwidth]{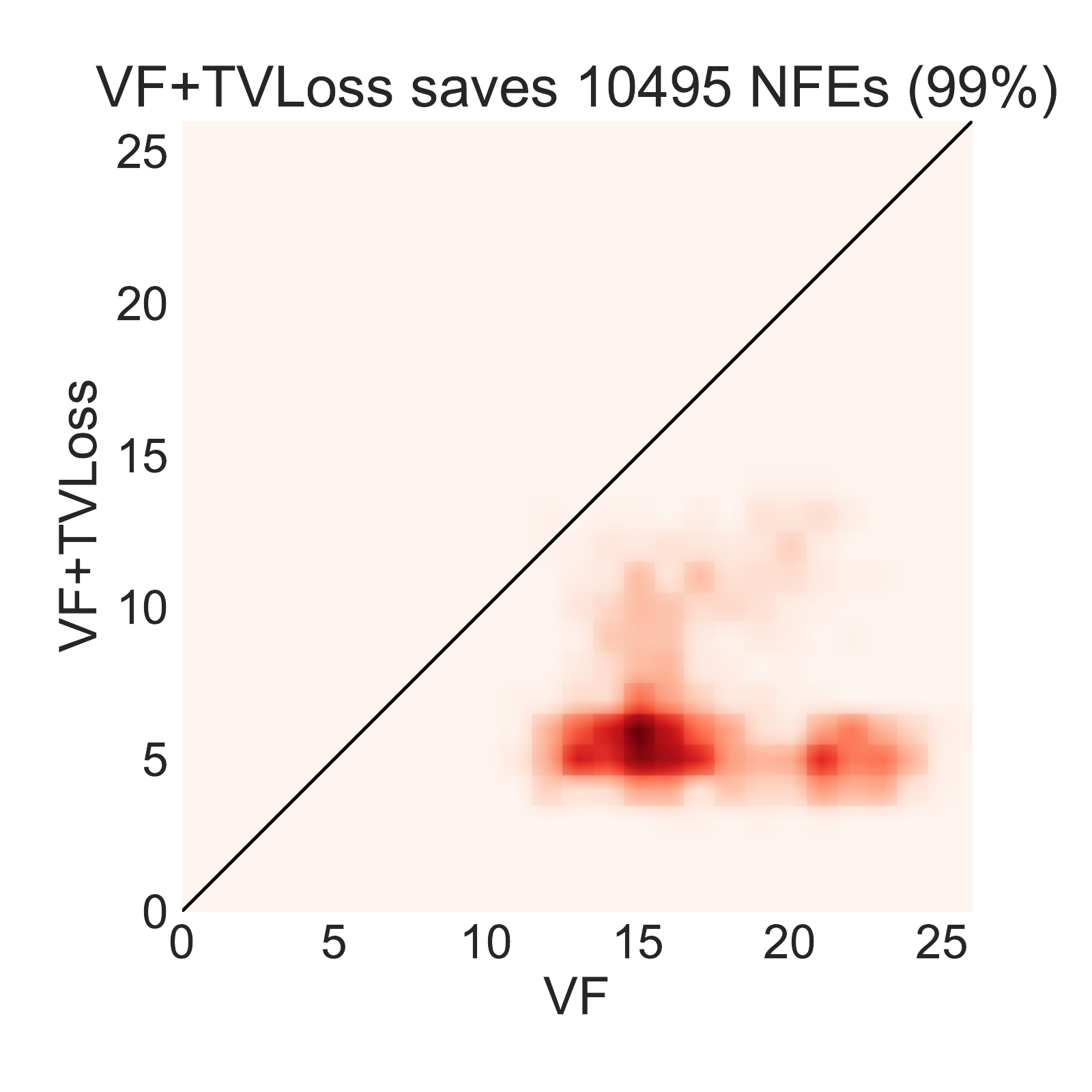}}
    \subfigure[Moons VF \vs SVFM]    {\includegraphics[width=0.3\textwidth]{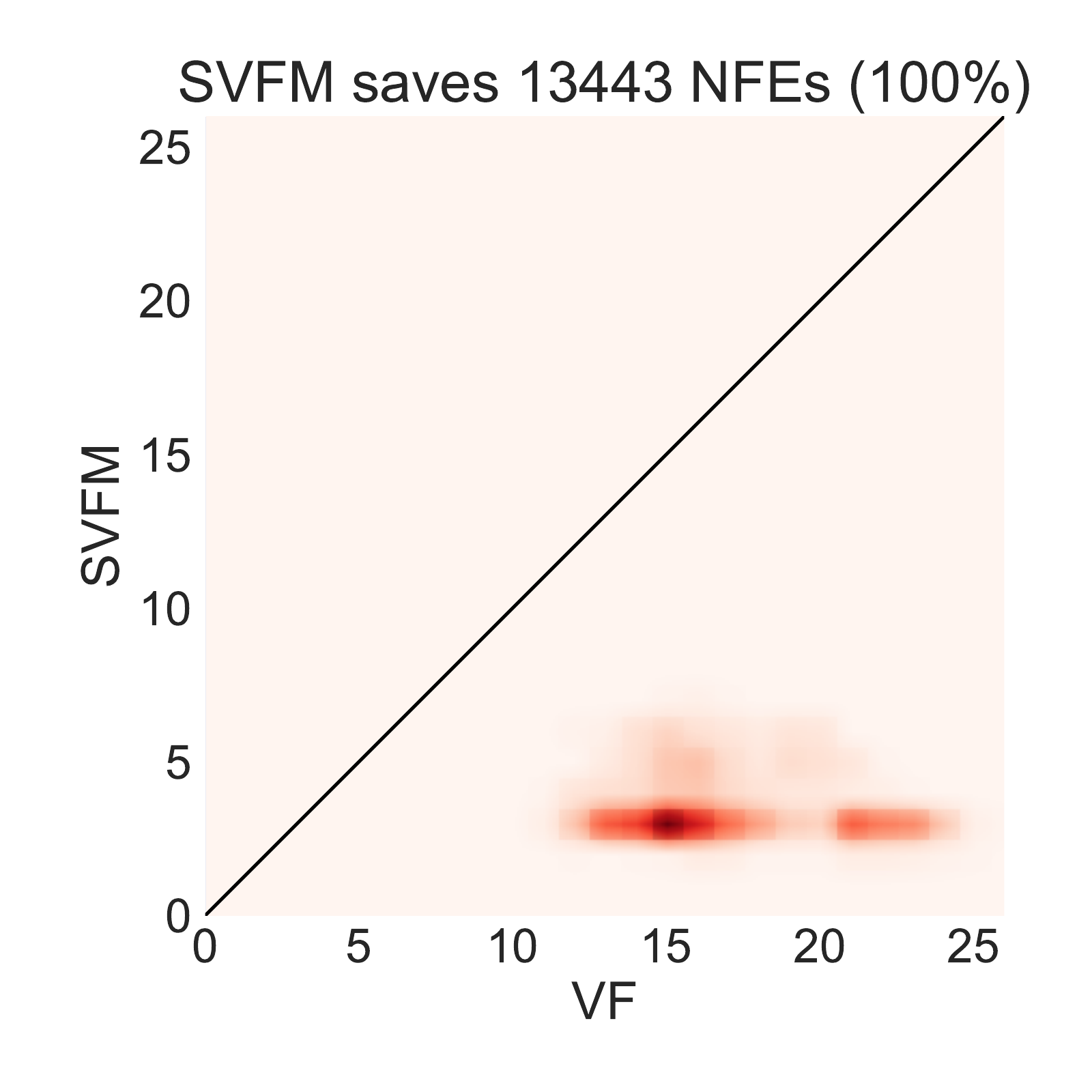}}
    \subfigure[Moons VF+TVL \vs SVFM]{\includegraphics[width=0.3\textwidth]{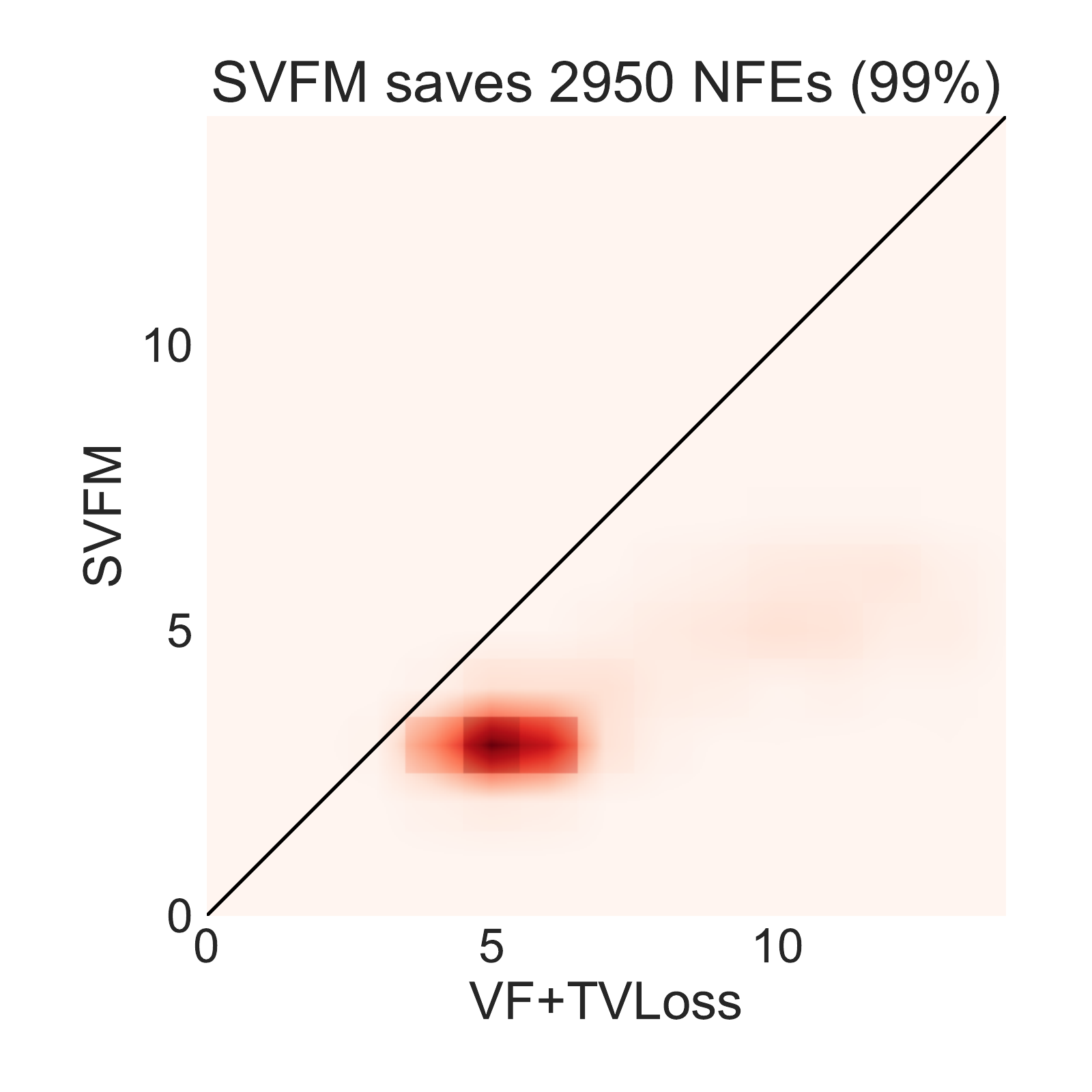}} 
    \\
    \subfigure[Circles VF \vs VF+TVL]  {\includegraphics[width=0.3\textwidth]{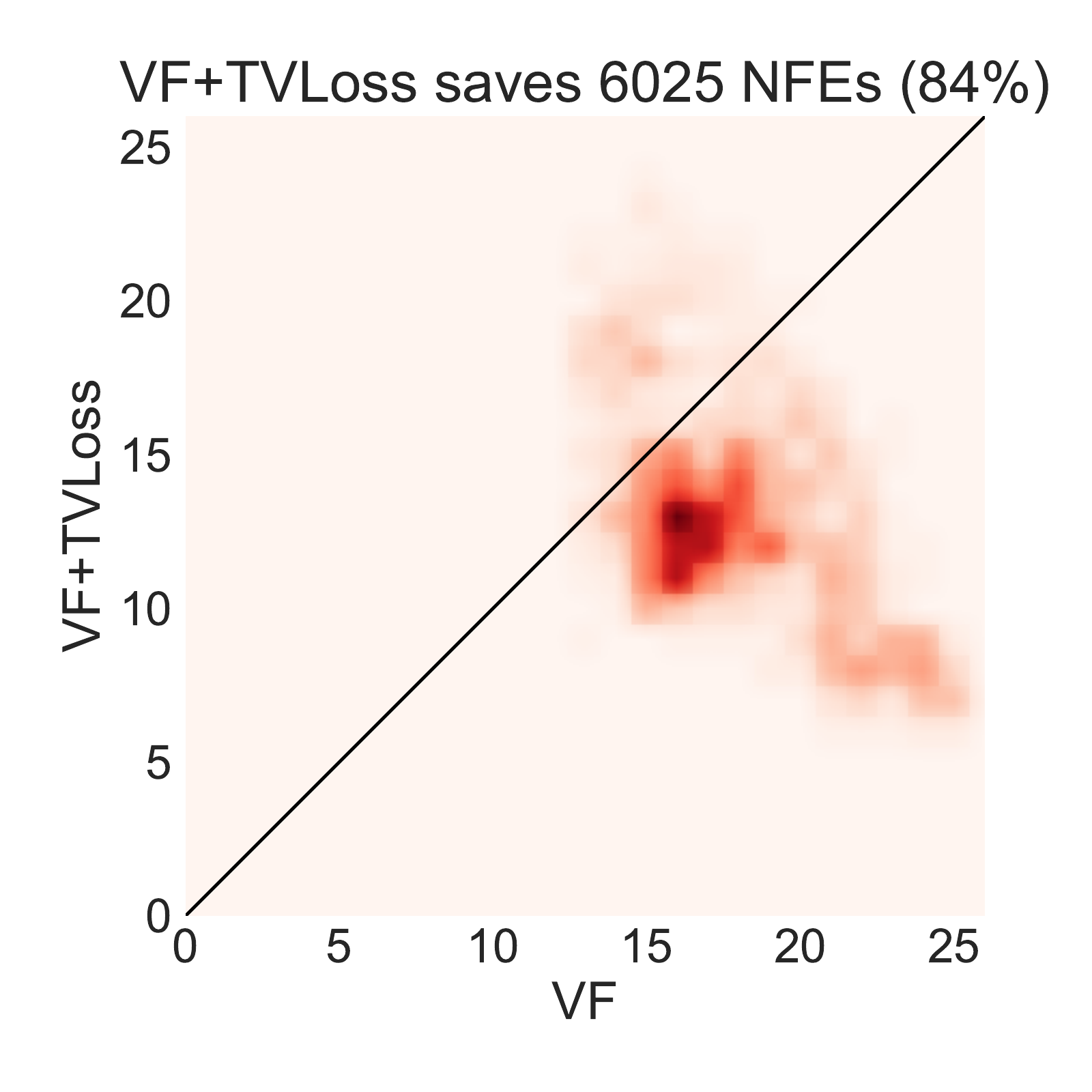}}
    \subfigure[Circles VF \vs SVFM]    {\includegraphics[width=0.3\textwidth]{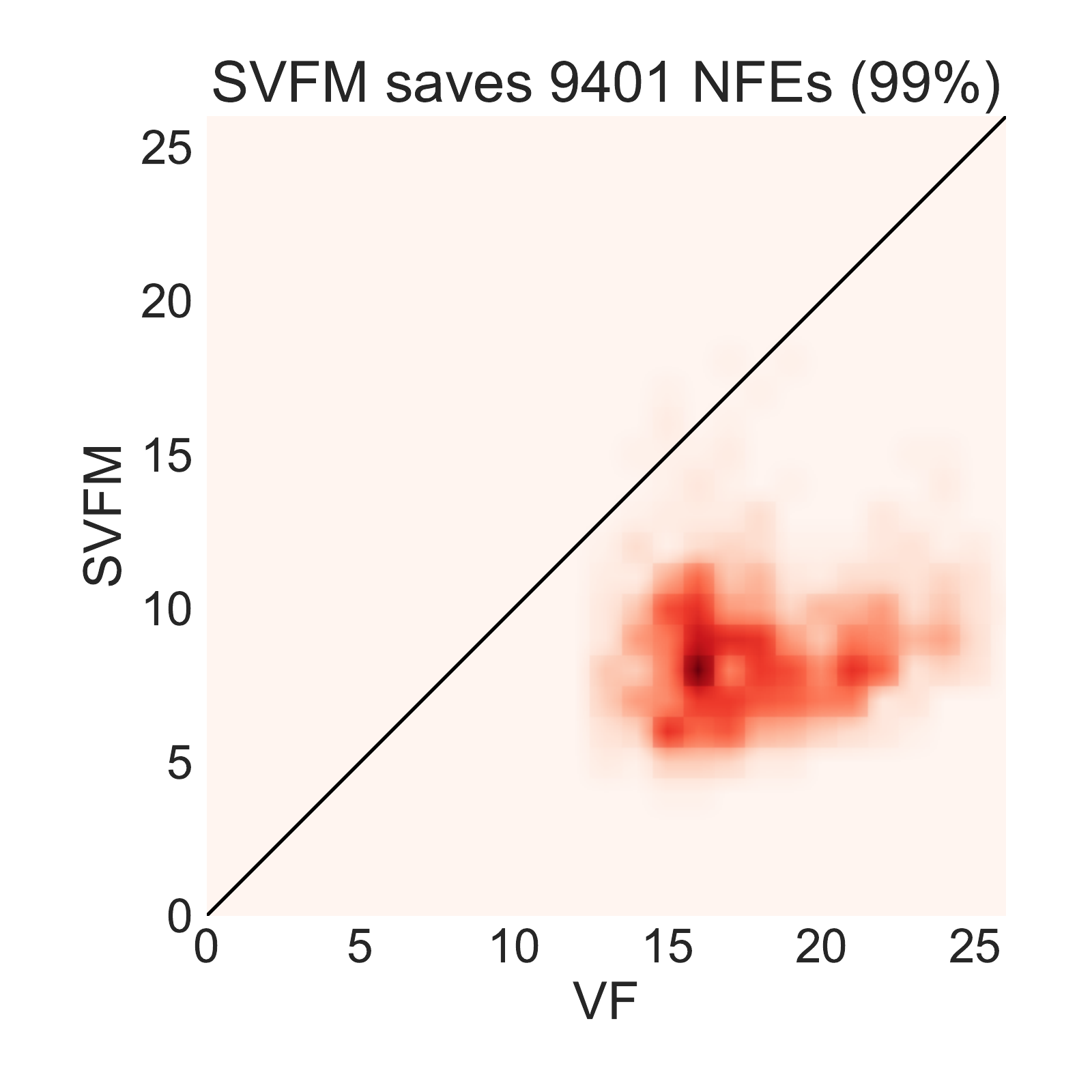}}
    \subfigure[Circles VF+TVL \vs SVFM]{\includegraphics[width=0.3\textwidth]{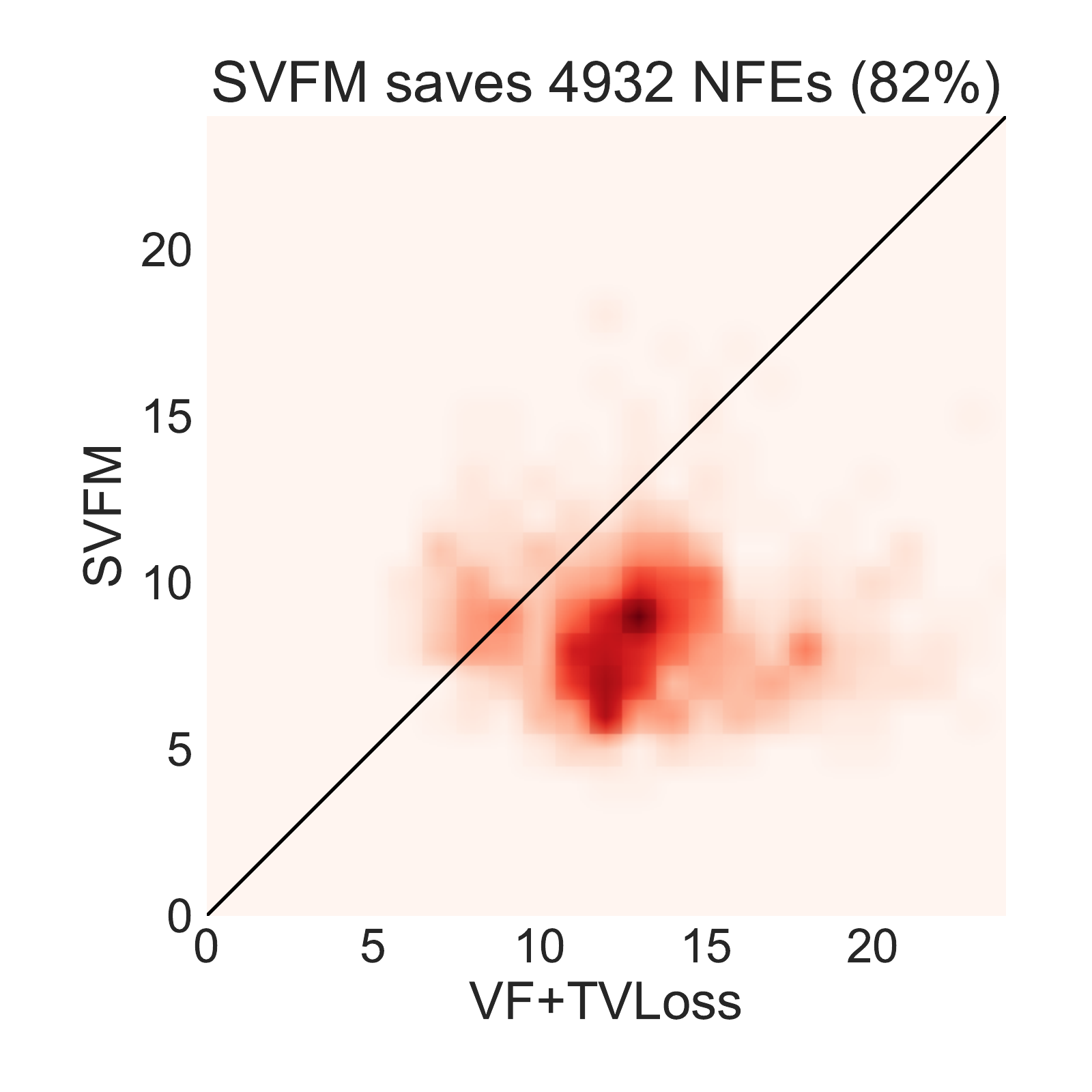}} 
    \\
    \subfigure[XOR VF \vs VF+TVL]  {\includegraphics[width=0.3\textwidth]{figures/xor/vf-vs-vf-myloss}}
    \subfigure[XOR VF \vs SVFM]    {\includegraphics[width=0.3\textwidth]{figures/xor/vf-vs-svfm}}
    \subfigure[XOR VF+TVL \vs SVFM]{\includegraphics[width=0.3\textwidth]{figures/xor/vf-myloss-vs-svfm}} 
    \caption{Per-instance NFE comparison densities on all illustrative experiments. The top, middle and lower rows arise from the moons, nested circles and XOR datasets respectively, and the left, middle and right columns compare VF \vs VF+TFLoss, VF \vs SVFM and VF+TFLoss \vs SVFM. In all cases significant savings are achieved with TFLosses and with SVFM models, with savings of $\approx$ 3,000-13,000 function evaluations over baseline approaches (or savings of 3-13 function evaluations on every instance, on average).}
    \label{fig:perinst_all}
\end{figure}

The figures in this section (\cref{fig:advection_illustration,fig:perinst_all}) are included to support and provide deeper intuition of the results given in the main paper. The captions provide self-contained descriptions of the main discussion points and conclusions that can be drawn from the pictures.

\end{appendices}

\end{document}